\documentclass[english,twoside,11pt]{article}

\usepackage{jmlr2e}
\usepackage{amsfonts}
\usepackage{amsmath}
\usepackage{multicol}
\usepackage{booktabs}

\usepackage{sidecap}
\usepackage{subfigure}
\usepackage{hyperref}
\hypersetup{
    colorlinks=false,
    pdfborder={0 0 0},
}

\firstpageno{1}
\iftrue
\fi

\newif\ifLongVersion
\LongVersionfalse

\usepackage{babel}

\begin{document}

\title{Knowledge Matters: Importance of Prior Information for Optimization}

\author{\name \c{C}a\u{g}lar G\"{u}l\c{c}ehre \email gulcehrc@iro.umontreal.ca\\
    \addr D\'epartement d'informatique et de recherche op\'erationnelle\\
    Universit\'e de Montr\'eal,  Montr\'eal, QC, Canada
    \AND
    \name Yoshua Bengio \email bengioy@iro.umontreal.ca \\
    \addr D\'epartement d'informatique et de recherche op\'erationnelle\\
    Universit\'e de Montr\'eal,  Montr\'eal, QC, Canada}

\editor{Not Assigned}

\maketitle

\begin{abstract}

We explore the effect of introducing prior information into the intermediate
level of deep supervised neural networks for a learning task on which all
the black-box state-of-the-art machine learning algorithms tested have failed to learn. We
motivate our work from the hypothesis that there is an optimization obstacle involved
in the nature of such tasks, and that humans learn useful intermediate
concepts from other individuals via a form of supervision or guidance using
a curriculum. The experiments we have conducted provide positive evidence in favor of this
hypothesis. In our experiments, a two-tiered MLP architecture is trained on
a dataset for which each image input contains three
sprites, and the binary target class is 1 if all three have the same shape.
Black-box machine learning algorithms only got chance on this task.
Standard deep supervised neural networks also failed. However, using
a particular structure and guiding the learner by providing intermediate
targets in the form of intermediate concepts (the presence of each object)
allows to nail the task. Much better than chance but imperfect results
are also obtained by exploring architecture and optimization variants,
pointing towards a difficult optimization task. We hypothesize that the learning
difficulty is due to the {\em composition} of
two highly non-linear tasks.  Our findings are also consistent with hypotheses
on cultural learning inspired by the observations of {\em effective} local minima (possibly
due to ill-conditioning and the training procedure not being able to escape what
appears like a local minimum).

\end{abstract}


\section{Introduction}

There is a recent emerging interest in different fields of science for
{\em cultural learning}~\citep{henrich2003evolution} and how groups of
individuals exchanging information can learn in ways superior to individual
learning. This is also witnessed by the emergence of new research fields such
as "Social Neuroscience". Learning from other agents in an environment by the means
of cultural transmission of knowledge with a peer-to-peer communication is 
an efficient and natural way of acquiring or propagating common knowledge. The most
popular belief on how the information is transmitted between individuals is that
bits of information are transmitted by small units, called memes, which 
share some characteristics of genes, such as self-replication, mutation and 
response to selective pressures~\citep{Dawkins-1976}.

This paper is based on the hypothesis (which is further elaborated in~\citet{Bengio-chapter-2013})
that human culture and the evolution of ideas have been crucial to counter an optimization
issue: this difficulty would otherwise make it
difficult for human brains to capture high level knowledge of the world without the
help of other educated humans.
In this paper machine learning experiments are used to investigate some elements of this hypothesis by
seeking answers for the following questions: are there machine learning tasks which are
intrinsically hard for a lone learning agent but that may become very easy
when intermediate concepts are provided by another agent as additional
intermediate learning cues, in the spirit of Curriculum Learning~\citep{Bengio+al-2009-small}?
What makes such learning tasks more difficult? Can
specific initial values of the neural network parameters yield success when
random initialization yield complete failure? Is it possible to verify that the problem being faced 
is an optimization problem or with a regularization problem?
These are the questions
discussed (if not completely addressed) here, which relate to the following 
broader question: how can humans (and potentially one day, machines) learn
complex concepts?

In this paper, results of different machine learning algorithms on an artificial learning task 
involving binary 64$\times$64 images are presented. In that task, each image in the dataset contains 3 Pentomino
tetris sprites (simple shapes). The task is to figure out if all the
sprites in the image are the same or if there are different sprite shapes in the
image. Several state-of-the-art machine learning algorithms have been tested and
none of them could perform better than a random predictor on the test
set. Nevertheless by providing hints about the intermediate concepts
(the presence and location of particular sprite classes), the problem can easily be solved
where the same-architecture neural network without the intermediate concepts
guidance fails. Surprisingly, our attempts at solving this problem with 
unsupervised pre-training algorithms failed solve this problem. 
However, with specific variations in the network architecture or training
procedure, it is found that one can make a big dent in the problem.
For showing the impact of intermediate level guidance,
we experimented with a two-tiered neural network,
with supervised pre-training of the first part to recognize 
the category of sprites independently of their 
orientation and scale, at different locations, 
while the second part learns from the output of the first
part and predicts the binary task of interest.

The objective of this paper is not to propose a novel learning algorithm or
architecture, but rather to refine our understanding of the learning difficulties
involved with composed tasks (here a logical formula composed with the detection
of object classes), in particular the training difficulties involved for
deep neural networks. The results also bring empirical evidence in favor of some
of the hypotheses from ~\citet{Bengio-chapter-2013}, discussed below, as well as 
introducing a particular form of curriculum learning~\citep{Bengio+al-2009-small}.

Building difficult AI problems has a long history in computer science. Specifically 
hard AI problems have been studied to create CAPTCHA's that are easy to solve for humans, but
hard to solve for machines \citep{von2003captcha}. In this paper we are investigating
a difficult problem for the off-the-shelf black-box machine learning algorithms.\footnote{You can
    access the source code of some experiments presented in that paper and their hyperparameters from
    here: \url{https://github.com/caglar/kmatters}}

\subsection{Curriculum Learning and Cultural Evolution Against Effective Local Minima}

What \citet{Bengio-chapter-2013} calls an {\bf effective local minimum} is a point
where iterative training stalls, either because of an actual local minimum
or because the optimization algorithm is unable (in reasonable time) to
find a descent path (e.g., because of serious ill-conditioning). In this paper, it is hypothesized
that some more abstract learning tasks such as those obtained by composing
simpler tasks are more likely to yield effective local minima for neural networks,
and are generally hard for general-purpose machine learning algorithms.

The idea that learning can be enhanced by guiding the learner through intermediate 
easier tasks is old, starting with animal training by {\em shaping}
~\citep{Skinner1958,Peterson2004,Krueger+Dayan-2009}.~\citet{Bengio+al-2009-small}
introduce a computational hypothesis related to a presumed issue with effective local minima
when directly learning the target task: the good solutions correspond to
hard-to-find-by-chance effective local minima, and intermediate tasks prepare the 
learner's internal configuration (parameters) in a way similar to continuation 
methods in global optimization (which go through a sequence
of intermediate optimization problems, starting with a convex one where local
minima are no issue, and gradually morphing into the target task of interest).

In a related vein,
\citet{Bengio-chapter-2013} makes the following inferences based on
experimental observations of deep learning and neural network learning:

\begin{description}
\item Point 1: Training deep architectures is easier when some hints are given about the function that the intermediate levels should compute \citep{Hinton06,WestonJ2008,Salakhutdinov2009,Bengio-2009}. 
{\em The experiments performed here expand in particular on this point.}

\item Point 2: It is much easier to train a neural network with supervision (where examples ar
provided to it of when a concept is present and when it is not present in a variety of examples) 
than to expect unsupervised learning to discover the concept (which may also happen but usually leads to poorer renditions of the concept). 
{\em The poor results obtained with unsupervised pre-training reinforce
that hypothesis}.

\item Point 3: Directly training all the layers of a deep network together not only makes it difficult to exploit all the extra modeling power of a deeper architecture but in many cases it actually yields worse results as the number of {\em required layers} is increased \citep{Larochelle-jmlr-toappear-2008,Erhan+al-2010}. {\em The experiments performed here also reinforce that hypothesis.}

\item Point 4: \citet{Erhan+al-2010} observed that no two training trajectories ended up in the same effective local minimum, out of hundreds of runs, even when comparing solutions as functions from input to output, rather than in parameter space (thus eliminating from the picture the presence of symmetries and multiple local minima due to relabeling and other reparametrizations). This suggests that the number of different effective local minima (even when considering them only in function space) must be huge.

\item Point 5: Unsupervised pre-training, which changes the initial
  conditions of the descent procedure, sometimes allows to reach
  substantially better effective local minima (in terms of generalization error!),
  and these better local minima do not appear to be reachable by chance
  alone \citep{Erhan+al-2010}. {\em The experiments performed here provide
    another piece of evidence in favor of the hypothesis that where random initialization
    can yield rather poor results, specifically targeted initialization can have a
    drastic impact, i.e., that effective local minima are not just numerous but that
    some small subset of them are much better and hard to reach by chance.}\footnote{Recent work
    showed that rather deep feedforward networks can be very successfully
    trained when large quantities of labeled data are
    available~\citep{Ciresan-2010,Glorot+al-AI-2011-small,Krizhevsky-2012}. Nonetheless,
    the experiments reported here suggest that it all depends on the task
    being considered, since even with very large quantities of labeled
    examples, the deep networks trained here were unsuccessful.}
\end{description}

Based on the above points, \citet{Bengio-chapter-2013} then proposed the following hypotheses 
regarding learning of high-level abstractions.
\begin{itemize}
\item \textbf{Optimization Hypothesis:} When it learns, a biological agent performs an approximate optimization with respect to some implicit objective function.
\item \textbf{Deep Abstractions Hypothesis:} Higher level abstractions represented in brains require deeper computations (involving the composition of more non-linearities).
\item \textbf{Local Descent Hypothesis:} The brain of a biological agent relies on approximate local descent and gradually improves itself while learning.
\item \textbf{Effective Local Minima Hypothesis:} The learning process of a single human learner (not helped by others) is limited by effective local minima.
\item \textbf{Deeper Harder Hypothesis:} Effective local minima are more likely to hamper learning as the required depth of the architecture increases.
\item \textbf{Abstractions Harder Hypothesis:} High-level abstractions are unlikely to be discovered by a single human learner by chance, because these abstractions
are represented by a deep subnetwork of the brain, which learns by local descent.
\item \textbf{Guided Learning Hypothesis:} A human brain can learn high level abstractions if guided by the signals produced by other agents that act as hints or
indirect supervision for these high-level abstractions.
\item \textbf{Memes Divide-and-Conquer Hypothesis:} Linguistic exchange, individual learning and the recombination of memes constitute an efficient evolutionary
recombination operator in the meme-space. This helps human learners to {\em collectively} build better internal representations of their environment,
including fairly high-level abstractions.
\end{itemize}

This paper is focused on ``\textit{Point 1}'' and testing the
``\textit{Guided Learning Hypothesis}'', using machine learning algorithms
to provide experimental evidence. The experiments performed also provide
evidence in favor of the ``\textit{Deeper Harder Hypothesis}'' and associated
``\textit{Abstractions Harder Hypothesis}''.
Machine Learning is still far beyond the
current capabilities of humans, and it is important to tackle the remaining obstacles
to approach AI. For this purpose, the question to be answered is why
tasks that humans learn effortlessly from very few examples, while machine
learning algorithms fail miserably?

\section{Culture and Optimization Difficulty}

As hypothesized in the ``\textit{Local Descent Hypothesis}'', human brains
would rely on a local approximate descent, just like a Multi-Layer
Perceptron trained by a gradient-based iterative optimization.
The main argument in favor of this hypothesis relies on the
biologically-grounded assumption that although firing patterns in the brain
change rapidly, synaptic strengths underlying these neural activities change only
gradually, making sure that behaviors are generally consistent across time.
If a learning algorithm is based on a form of local (e.g. gradient-based) descent,
it can be sensitive to effective local minima ~\citep{Bengio-chapter-2013}.

When one trains a neural network, at some point in the training phase the
evaluation of error seems to saturate, even if new examples are 
introduced. In particular~\citet{Erhan+al-2010} find that early examples
have a much larger weight in the final solution. It looks like the learner is
stuck in or near a local minimum. But since it is difficult to verify
if this is near a true local minimum or simply an effect of strong ill-conditioning,
we call such a ``stuck'' configuration an {\em effective local minimum},
whose definition depends not just on the optimization objective but also on the
limitations of the optimization algorithm.

\citet{Erhan+al-2010} highlighted both the issue of effective local minima
and a regularization effect when initializing a deep network with unsupervised
pre-training. Interestingly, as the network gets deeper the difficulty due to effective local
minima seems to be get more pronounced. That might be because of the number of effective local
minima increases (more like an actual local minima issue),
or maybe because the good ones are harder to reach (more like an ill-conditioning
issue) and more work will be needed to clarify this question.

As a result of Point 4 we hypothesize that it is very difficult for an individual's
brain to discover some higher level abstractions by chance only. As mentioned in the ``\textit{Guided
Learning Hypothesis}'' humans get hints from other humans and learn high-level
concepts with the guidance of other humans\footnote{But some high-level concepts
may also be hardwired in the brain, as assumed in the universal grammar hypothesis
~\citep{montague1970universal}, or in nature vs nurture discussions in cognitive science.}.
Curriculum learning  \citep{Bengio+al-2009} and 
incremental learning \citep{solomonoff1989system},
are examples of this. This is done by properly choosing the sequence of examples
seen by the learner, where simpler examples are introduced first and 
more complex examples shown when the learner is ready for them. One of the hypothesis
on why curriculum works states that curriculum learning acts as a continuation method that allows
one to discover a good minimum, by first finding a good minimum of a smoother error function. Recent
experiments on human subjects also indicates that humans {\em teach}
by using a curriculum strategy \citep{Khan+Zhu+Mutlu-2011}.

Some parts of the human brain are known to have a hierarchical organization
(i.e. visual cortex) consistent with the deep architecture studied in machine
learning papers. As we go from the sensory level to higher levels of the
visual cortex, we find higher level areas corresponding to more abstract
concepts. This is consistent with the {\em Deep Abstractions Hypothesis}.

Training neural networks and machine learning algorithms by decomposing
the learning task into sub-tasks and exploiting prior information about the
task is well-established and in fact constitutes the main approach to
solving industrial problems with machine learning.
The contribution of this paper is rather on rendering explicit the effective local minima
issue and providing evidence on the type of problems for
which this difficulty arises.  This prior information and hints given
to the learner can be viewed as inductive bias for a particular task,
an important ingredient to obtain a good generalization error \citep{mitchell1980need}.
An interesting earlier finding in that line of research was done
with Explanation Based Neural Networks (EBNN) in which a neural network
transfers knowledge across multiple learning tasks. An EBNN uses previously
learned domain knowledge as an initialization or search bias (i.e. to
constrain the learner in the parameter space) ~\citep{Sullivan-96-integrating,mitchell1993explanation}.

Another related work in machine learning is mainly focused on reinforcement
learning algorithms, based on incorporating prior knowledge in terms of logical rules to the learning
algorithm as a prior knowledge to speed up and bias learning ~\citep{kunapuliadviceptron,towell1994knowledge}.

As discussed in ``\textit{Memes Divide and Conquer Hypothesis}`` societies can be viewed 
as a distributed computational processing systems. In civilized societies knowledge is
distributed across different individuals, this yields a space efficiency. Moreover computation,
i.e. each individual can specialize on a particular task/topic, is also divided across the individuals
in the society and hence this will yield a computational efficiency. Considering the limitations of the
human brain, the whole processing can not be done just by a single agent in an efficient manner. A recent
study in paleoantropology states that there is a substantial decline in endocranial volume of the brain
in the last 30000 years \cite{henneberg1988decrease}. The volume of the brain shrunk to 1241 ml from
1502 ml \citep{henneberg1993trends}. One of the hypothesis on the reduction of the volume of skull
claims that, decline in the volume of the brain might be related to the functional changes in brain
that arose as a result of cultural development and emergence of societies given that this time period
overlaps with the transition from hunter-gatherer lifestyle to agricultural societies.

\section{Experimental Setup}

Some tasks, which seem reasonably easy for humans to learn\footnote{keeping in mind that
humans can exploit prior knowledge, either from previous learning or innate knowledge.},
are nonetheless appearing almost impossible to learn for current generic state-of-art 
machine learning algorithms.

Here we study more closely such a task, which becomes learnable if one provides hints to the learner
about appropriate intermediate concepts. Interestingly, the task we used in our experiments
is not only hard for deep neural networks but also for non-parametric machine learning
algorithms such as SVM's, boosting and decision trees.

The result of the experiments for varying size of dataset with several off-the-shelf black box
machine learning algorithms and some popular deep learning algorithms are provided in Table
\ref{tab:results_cmp}. The detailed explanations about the algorithms and the hyperparameters
used for those algorithms are given in the Appendix Section \ref{sec:setup}. We also
provide some explanations about the methodologies conducted for the experiments at Section \ref{sec:learn_eva}.

\begin{figure}[htbp!]
\centering
\mbox{
 \subfigure[sprites, not all same type]{
 \includegraphics[scale=0.4]{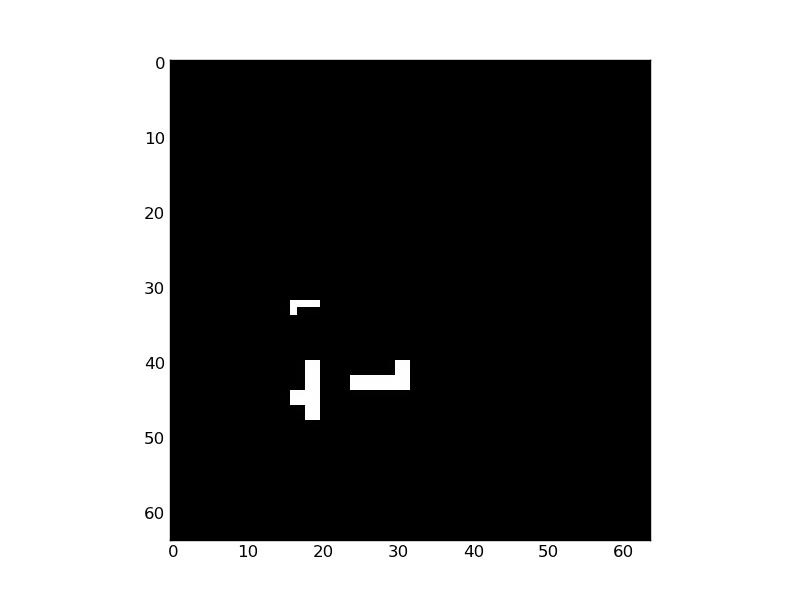}
 \label{fig:ex_img_different_objects}
 }
 \hspace*{-20mm}
 \subfigure[sprites, all of same type]{
 \includegraphics[scale=0.4]{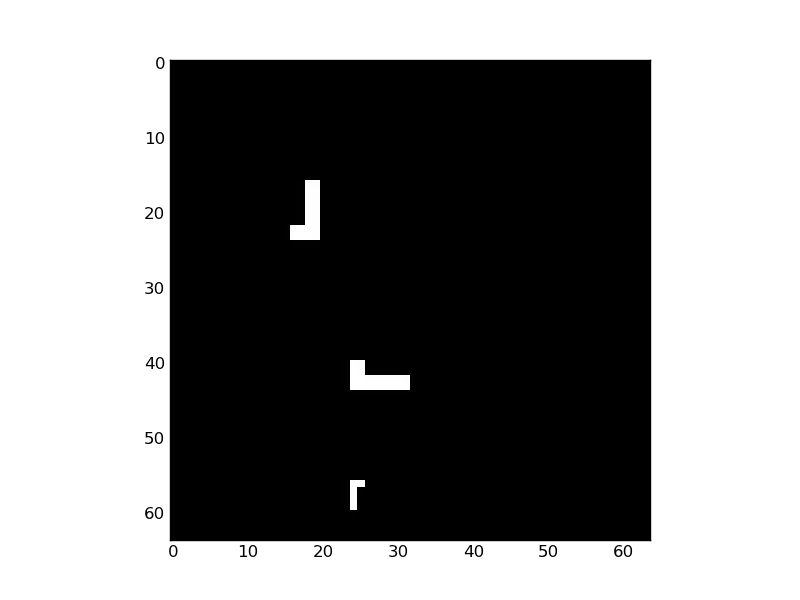}
 \label{fig:ex_img_same_objects}
 }
}
\caption{Left (a): An example image from the dataset which has {\em a different sprite type} in it. 
    Right (b): An example image from the dataset that has only one type of Pentomino object in it, 
    but with different orientations and scales.}
\label{fig:pento_same_diff}
\end{figure}

\subsection{Pentomino Dataset}

In order to test our hypothesis, an artificial dataset for
object recognition using 64$\times$64 binary images is designed\footnote{The source code for
the script that generates the artificial Pentomino datasets (Arcade-Universe) is available at:
\url{https://github.com/caglar/Arcade-Universe}. This implementation is based on
Olivier Breuleux's bugland dataset generator.}. If the task is two tiered (i.e., with guidance provided),
the task in the first part is to recognize and locate each Pentomino object class\footnote{~A human
learner does not seem to need to be taught the shape categories of each Pentomino
sprite in order to solve the task. On the other hand, humans have lots of previously
learned knowledge about the notion of shape and how central it is in defining
categories.} in the image. The second part/final binary classification task is to figure out
if all the Pentominos in the image are of the same shape class or not. If a neural network
learned to detect the categories of each object at each location in an image, the remaining
task becomes an XOR-like operation between the detected object categories. The types of Pentomino 
objects that is used for generating the dataset are as follows:

\ifLongVersion
\else

Pentomino sprites N, P, F, Y, J, and Q, along with the
Pentomino N2 sprite (mirror of ``Pentomino N'' sprite),
the Pentomino F2 sprite (mirror of ``Pentomino F'' sprite),
and the Pentomino Y2 sprite (mirror of ``Pentomino Y'' sprite).
\fi

\begin{figure}[htbp!]
 \centering
 \vspace*{-1mm}
 \includegraphics[scale=0.8]{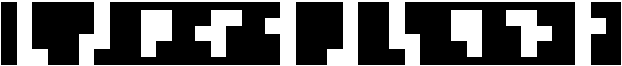}
 \vspace*{-2mm}
 \caption{Different classes of Pentomino shapes used in our dataset.}
 \label{fig:pento_classes}
\end{figure}

As shown in Figures \ref{fig:ex_img_different_objects} and
\ref{fig:ex_img_same_objects}, the synthesized images are fairly simple and do not have
any texture. Foreground pixels are ``1'' and background pixels are
``0''. Images of the training and test sets are generated iid. 
For notational convenience, assume that the
domain of raw input images is $X$, the set of sprites is $S$, the set of
intermediate object categories is $Y$ for each possible location in the image
and the set of final binary task outcomes is $Z=\{0,1\}$. 
Two different types of rigid body transformation is performed: sprite
rotation $rot(X, \gamma)$ where $\Gamma=\{\gamma \colon (\gamma=90 \times
\phi)~{\rm \land}~[(\phi \in \mathbb{N}), (0 \leq \phi \leq 3)]\}$ and scaling
$scale(X, \alpha)$ where $\alpha \in \{1,2\}$ is the scaling factor.
The data generating procedure is summarized below.

\begin{description}

 \item Sprite transformations: Before placing the sprites in an empty image,
   for each image $x \in X$, a value for $z \in Z$ is randomly sampled which is to have (or not) the
   same three sprite shapes in the image. Conditioned on the constraint given by $z$,
   three sprites are randomly selected $s_{ij}$ from $S$ without replacement. Using a uniform
   probability distribution over all possible scales, a scale is chosen and accordingly
   each sprite image is scaled. Then rotate each sprite is randomly rotated by a multiple of 90 degrees.

 \item Sprite placement: Upon completion of sprite transformations, a 
 64$\times$64 uniform grid is generated which is divided into 8$\times$8 blocks,
 each block being of size 8$\times$8 pixels, and randomly select three different blocks
 from the 64=8$\times$8 on the grid and place the
 transformed objects into {\em different} blocks (so they cannot overlap, by construction).

\end{description}

Each sprite is centered in the block in which it is located. Thus there is no
object translation inside the blocks. The only translation invariance is due
to the location of the block inside the image.
 
A Pentomino sprite is guaranteed to not overflow the block in which it 
is located, and there are no collisions or overlaps between sprites, making
the task simpler. The largest possible Pentomino sprite can be fit into an 8$\times$4 
mask.

\subsection{Learning Algorithms Evaluated}
\label{sec:learn_eva}
Initially the models are cross-validated by using 5-fold cross-validation.
With 40,000 examples, this gives 32,000 examples for training and 
8,000 examples for testing.
For neural network algorithms, stochastic gradient
descent (SGD) is used for training. The following standard learning
algorithms were first evaluated: decision trees, SVMs with Gaussian kernel,
ordinary fully-connected Multi-Layer Perceptrons, Random Forests,
k-Nearest Neighbors, Convolutional Neural Networks, and Stacked
Denoising Auto-Encoders with supervised fine-tuning. More details
of the configurations and hyper-parameters for each of them
are given in Appendix Section~\ref{sec:setup}. The only better than chance
results were obtained with variations of the Structured Multi-Layer Perceptron
described below.

\subsubsection{Structured Multi-Layer Perceptron (SMLP)}

\begin{figure}[htbp!]
 \centering
 \vspace*{-\parskip}
\includegraphics[width=\linewidth]{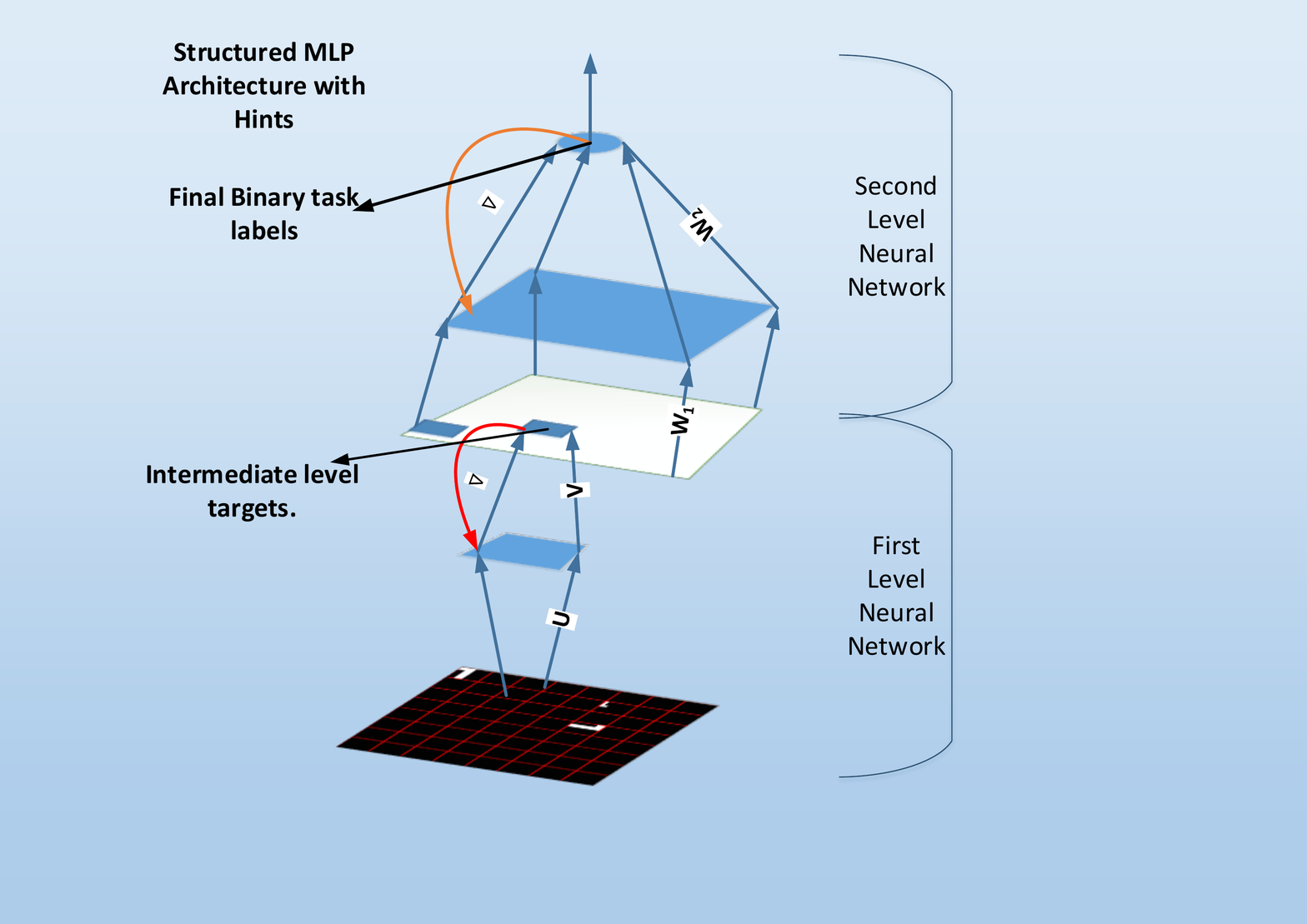}
  \vspace*{-0.2cm}
 \caption{Structured MLP architecture, used with hints (trained in two phases,
first P1NN, bottom two layers, then P2NN, top two layers). In SMLP-hints, P1NN is 
trained on each 8x8 patch extracted from the image and the softmax output
probabilities of all 64 patches are concatenated into a 64$\times$11 vector
that forms the input of P2NN. Only $U$ and $V$ are learned in the P1NN and its output on each 
patch is fed into P2NN. 
The first level and the second level neural networks are trained separately, not jointly.
 }\label{fig:net_arch_hints}
\end{figure}

The neural network architecture that is used to solve this task is called the SMLP (Structured Multi-Layer Perceptron),
a deep neural network with two parts as illustrated in Figure~\ref{fig:net_arch_hints} and \ref{fig:net_arch_nohints}:

The lower part, P1NN (\emph{Part 1 Neural Network}, as it is called in the rest of the paper),
has shared weights and local connectivity, with one identical 
MLP instance of the P1NN for each patch of the image,
and typically an 11-element output vector per patch (unless otherwise noted). 
The idea is that these 11 outputs per patch could represent
the detection of the sprite shape category (or the absence of sprite in the patch).
The upper part, P2NN ({\em Part 2 Neural Network}) is a fully connected one hidden layer MLP
that takes the concatenation of the outputs of all patch-wise P1NNs as input. Note that
the first layer of P1NN is similar to a convolutional layer
but where the stride equals the kernel size, so that windows do not overlap, i.e., P1NN can be
decomposed into separate networks sharing the same parameters but applied
on different patches of the input image, so that each network can actually be trained
patch-wise in the case where a target is provided for the P1NN outputs. The P1NN output
for patch $\mathbf{p_i}$ which is extracted from the image $\mathbf{x}$ is computed as follows:

\begin{equation}
    f_{\theta}(\mathbf{p_i}) = g_2(V g_1(U \mathbf{p_i} + \mathbf{b}) + \mathbf{c})
\end{equation}

where $\mathbf{p_i} \in R^d$ is the input patch/receptive field extracted from location $i$ of a single image.
$U \in R^{d_h \times d}$ is the weight matrix for the first layer of P1NN and 
$\mathbf{b} \in R^d_h$ is the vector of biases for the first layer of P1NN.
$g_1(\cdot)$ is the activation function of the first layer and
$g_2(\cdot)$ is the activation function of the second layer. In many of the experiments, best
results were obtained with $g_1(\cdot)$ 
a rectifying non-linearity (a.k.a. as RELU), which is $max(0,~X)$
~\citep{Jarrett-ICCV2009-small,Hinton2010,Glorot+al-AI-2011-small,Krizhevsky-2012}. 
$V \in R^{d_h \times d_o}$ is the second layer's weights matrix, such that and 
$\mathbf{c} \in R_{d_o}$ are the biases of the second layer of the P1NN, with 
$d_o$ expected to be smaller than $d_h$.

In this way, $g_1(U \mathbf{p_i} + \mathbf{b})$ is an overcomplete representation of the input patch
that can potentially represent all the possible Pentomino shapes for all factors of variations in the
patch (rotation, scaling and Pentomino shape type). On the other hand, when trained with
hints, $f_\theta(\mathbf{p_i})$ is expected to be the lower dimensional representation of a Pentomino
shape category invariant to scaling and rotation in the given patch.

In the experiments with SMLP trained with hints (targets at the output of P1NN),
%
the P1NN is expected to perform classification of each 8$\times$8 non-overlapping patches
of the original 64$\times$64 input image without having any prior knowledge of whether
that specific patch contains a Pentomino shape or not. P1NN in SMLP without hints just outputs the
local activations for each patch, and gradients on $f_\theta({\mathbf p_i})$ are backpropagated
from the upper layers. In both cases P1NN produces the input representation  for the Part 2 Neural Net (P2NN). 
Thus the input representation of P2NN is the concatenated output of P1NN across all the 64 patch locations:

$\mathbf{h_o} = [f_{\theta}(\mathbf{p_0}), ..., f_{\theta}(\mathbf{p_i}), ...,
       f_{\theta}(\mathbf{p_{N}}))]$ where $N$ is the number of patches and the $h_o \in R^{d_i}, d_i = d_o \times N$.
       $\mathbf{h_o}$ is the concatenated output of the P1NN at each patch.

There is a standardization layer on top of the output of P1NN that centers the activations and
performs divisive normalization by dividing  by the standard deviation over a minibatch
of the activations of that
layer. We denote the standardization function $z(\cdot)$. Standardization makes use of the
mean and standard deviation computed for each hidden unit such that each hidden unit of 
$\mathbf{h_o}$ will have 0 activation and unit standard deviation on average
over the minibatch. $X$ is the set of
pentomino images in the minibatch, where $X \in R^{d_{in} \times N}$ is a matrix with $N$ images.
$h_o^{(i)}(\mathbf{x_j})$ is the vector of activations of the $i$-th hidden unit of
hidden layer $h_o(\mathbf{x_j})$ for the $j$-th example, with $x_j \in X$.

\begin{equation}
\label{eq:mu}
\mu_{h_o^{(i)}} = \frac{1}{N}\sum_{\mathbf{x_j} \in X} h_o^{(i)}(\mathbf{x_j})
\end{equation}

\begin{equation}
\label{eq:sigma}
\sigma_{h_o^{(i)}} = \sqrt{\frac{\sum_j^N (h_o^{(i)}(\mathbf{x_j})-\mu_{h_o^{(i)}})^2}{N} +
    \epsilon}
\end{equation}
\begin{equation}
\label{eq:std}
z(h_o^{(i)}(\mathbf{x_j}))=\frac{h_o^{(i)}(\mathbf{x_j}) - \mu_{h_o^{(i)}}}{\max
    (\sigma_{h_o^{(i)}}, \epsilon)}
\end{equation}

where $\epsilon$ is a very small constant, that is used to prevent numerical underflows in the
standard deviation. P1NN is trained on each 8$\times$8 patches extracted from the image.

$\mathbf{h_o}$ is standardized for each training and test sample separately. Different values of $\epsilon$ were used for SMLP-hints and SMLP-nohints.

The concatenated output of P1NN is fed as an input to the P2NN. P2NN is a feedforward MLP with a
sigmoid output layer using a single RELU hidden layer. The task of P2NN is to perform a nonlinear logical
operation on the representation provided at the output of P1NN.

\subsubsection{Structured Multi Layer Perceptron Trained with Hints (SMLP-hints)}

The SMLP-hints architecture exploits a hint about the presence and category
of Pentomino objects, specifying a semantics for the P1NN outputs. P1NN is trained with the
intermediate target $Y$, specifying the type of Pentomino sprite shape present (if any) at each of
the 64 patches (8$\times$8 non-overlapping blocks) of the image. Because a possible answer at a given
location can be ``none of the object types'' i.e., an empty patch, $y_p$ (for patch $p$) can take
one of the 11 possible values, 1 for rejection and the rest is for the Pentomino shape classes,
illustrated in Figure \ref{fig:pento_classes}:

\begin{equation*}
y_p =
    \begin{cases}
        0 & \text{if  patch~} p \text{~is empty~} \\
        s \in S & \text{if the patch~} p \text{~contains a Pentomino sprite~}.
    \end{cases}
\end{equation*}

A similar task has been studied by \citet{fleuret2011comparing} (at SI appendix Problem 17),
who compared the performance of humans vs computers.

The SMLP-hints architecture takes advantage of dividing the task into two subtasks during training
with prior information about intermediate-level relevant factors. 
Because the sum of the training losses decomposes into the loss on each patch, the P1NN can be
pre-trained patch-wise. Each patch-specific component of the P1NN is
a fully connected MLP with 8$\times$8 inputs and 11 outputs with a softmax output layer.
SMLP-hints uses the the standardization given in Equation \ref{eq:sigma} but with $\epsilon=0$.

\begin{figure}[htp]
\centering{
\includegraphics[scale=0.6]{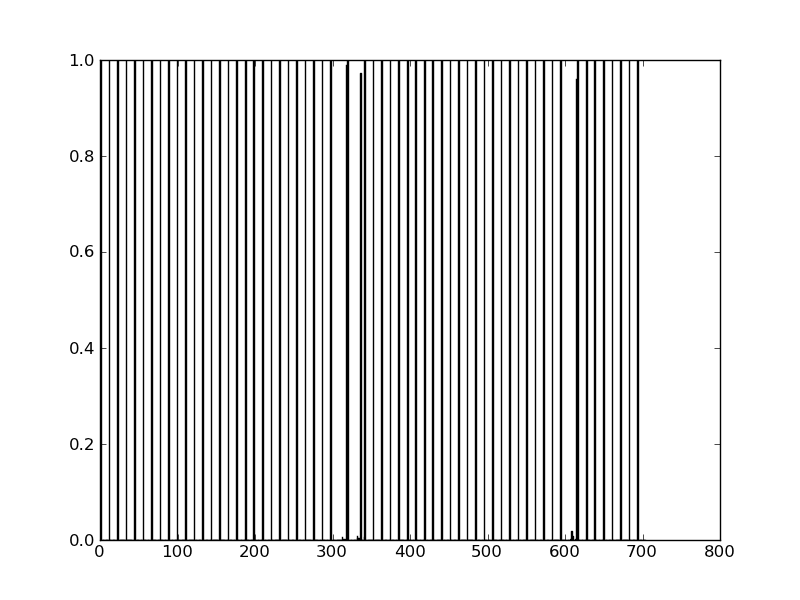}
}
\caption{Bar chart of concatenated softmax output activations $\mathbf{h_o}$ of P1NN (11$\times$64=704 outputs) in SMLP-hints before standardization, for a selected example. There are
    very large spikes at each location for one of the possible 11 outcome (1 of K representation).}
\label{fig:ikgnn_unstd_act_bar}
\end{figure}

\begin{figure}[htp]
\centering{
\includegraphics[scale=0.6]{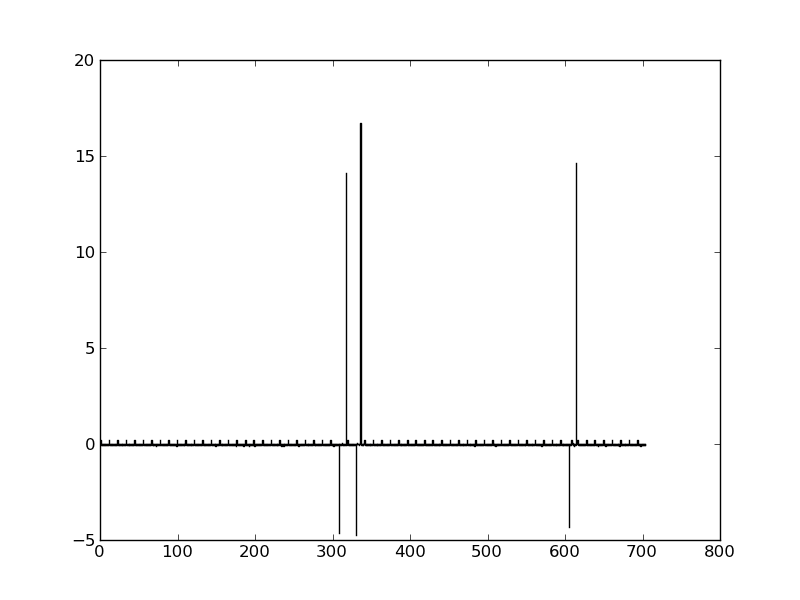}
}
\caption{Softmax output activations $\mathbf{h_o}$ of P1NN  at SMLP-hints before standardization.
    There are positive spiked outputs at the locations where there is a Pentomino shape. 
    Positive and negative spikes arise because most of the outputs are near an average value.
    Activations are higher at the locations where there is a pentomino shape.}
\label{fig:ikgnn_std_act_bar}
\end{figure}

The standardization is a crucial step for training the SMLP on the Pentomino dataset, and yields
much sparser outputs, as seen on
Figures \ref{fig:ikgnn_unstd_act_bar} and \ref{fig:ikgnn_std_act_bar}. If the standardization is
not used, even SMLP-hints could not solve the Pentomino task. 
In general, the standardization step dampens the small activations and augments larger
ones(reducing the noise). Centering the activations of each feature detector in a neural network has been
studied in \citep{raiko2012deep} and \citep{vatanen2013pushing}. They proposed that transforming the outputs
of each hidden neuron in a multi-layer perceptron network to have zero output and zero slope on average makes
first order optimization methods closer to the second order techniques.

By default, the SMLP uses rectifier hidden units as activation function,  we
found a significant boost by using rectification compared to hyperbolic
tangent and sigmoid activation functions. The P1NN has a highly overcomplete architecture
with 1024 hidden units per patch, and L1 and L2 weight decay regularization 
coefficients on the weights (not the biases) are respectively 1e-6 and 1e-5. The learning rate for
the P1NN is 0.75. 1 training epoch was enough for the P1NN to learn the features of 
Pentomino shapes perfectly on the 40000 training examples. The P2NN has 2048 hidden units.
L1 and L2 penalty coefficients for the P2NN are 1e-6, and the learning rate is 0.1. These were 
selected by trial and error based on validation set error. Both P1NN (for each patch)
and P2NN are fully-connected neural networks, even though P1NN globally is a special
kind of convolutional neural network.

\begin{figure}[htbp!]
 \centering
 \vspace*{-\parskip}
\includegraphics[scale=1.8]{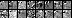}
 \caption{Filters of Structured MLP architecture, trained with hints on 40k examples.}
 \label{fig:smlp_hints_filters_patches}
\end{figure}

Filters of the first layer of SMLP are shown in Figure \ref{fig:smlp_hints_filters_patches}.
These are the examples of the filters obtained with the SLMP-hints trained with 40k examples,
whose results are given in Table \ref{tab:results_cmp}. Those filters look very noisy but they work perfectly on the Pentomino task.

\subsubsection{Deep and Structured Supervised MLP without Hints (SMLP-nohints)}

SMLP-nohints uses the same connectivity pattern (and deep architecture)
that is also used in the SMLP-hints architecture, but without using the intermediate targets ($Y$).
It directly predicts the final outcome of the task ($Z$),
using the same number of hidden units, the same connectivity and the same
activation function for the hidden units as SMLP-hints. 
120 hyperparameter values have been evaluated
by randomly selecting the number of hidden units from $[64, 128, 256, 512, 1024, 1200, 2048]$ and randomly sampling 20 
learning rates uniformly in the log-domain within the interval of $[0.008, 0.8]$. Two fully connected
hidden layers with 1024 hidden units (same as P1NN) per patch is used and 2048 (same as P2NN) for the last hidden layer,
with twenty training epochs. For this network the best results are obtained with a learning rate
of 0.05.\footnote{The source code of the structured MLP is available at the github repository:
    \url{https://github.com/caglar/structured\_mlp}}

\begin{figure}[htbp!]
 \centering
 \vspace*{-\parskip}
 \centering
\includegraphics[width=\linewidth]{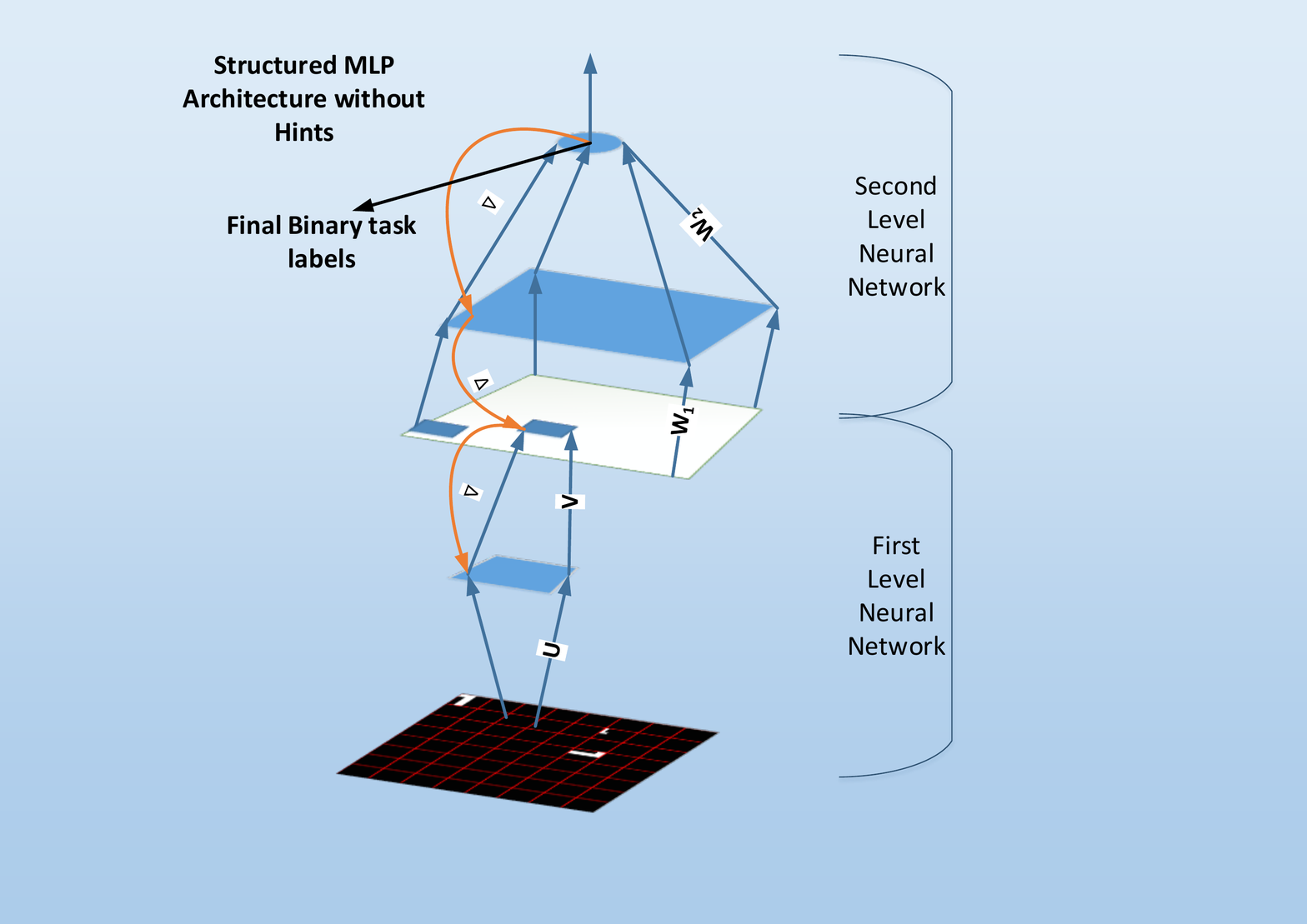}
  \vspace*{-0.2cm}
 \caption{Structured MLP architecture, used without hints (SMLP-nohints). 
It is the same architecture as SMLP-hints (Figure~\ref{fig:net_arch_hints}) but with
both parts (P1NN and P2NN) trained jointly with respect to the final binary
classification task.}
 \label{fig:net_arch_nohints}
\end{figure}

\begin{figure}[htbp!]
 \centering
 \vspace*{-\parskip}
\includegraphics[scale=1.8]{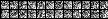}
 \caption{First layer filters learned by 
   the Structured MLP architecture, trained without using hints on 447600 examples
     with online SGD and a sigmoid intermediate layer activation.}
 \label{fig:osmlp_nohints_filters_subset}
\end{figure}

We chose to experiment with various SMLP-nohint architectures and optimization procedures, trying
unsuccessfully to achieve as good results with SMLP-nohint as with SMLP-hints.

\paragraph{Rectifier Non-Linearity}
A rectifier nonlinearity is used for the activations of MLP hidden layers. We observed that using
piecewise linear nonlinearity activation function such as the rectifier can make the optimization more
tractable.

\paragraph{Intermediate Layer}

The output of the P1NN is considered as an intermediate layer of the SMLP. 
For the SMLP-hints, only softmax output activations have been tried at the intermediate layer, and that sufficed
to learn the task. Since things did not work nearly as well with the SMLP-nohints, 
several different activation functions have been tried: softmax$(\cdot)$, tanh$(\cdot)$, sigmoid$(\cdot)$ and linear activation functions.

\paragraph{Standardization Layer}

Normalization at the last layer of the convolutional neural networks has been used occasionaly to
encourage the competition between the hidden units. \citep{Jarrett-ICCV2009} used a local contrast
normalization layer in their architecture which performs subtractive and divisive normalization.
A local contrast normalization layer enforces a local competition between adjacent features in the
feature map and between features at the same spatial location in different feature maps. Similarly
\citep{Krizhevsky-2012} observed that using a local response layer that enjoys the benefit of using
local normalization scheme aids generalization.

Standardization has been observed to be crucial for both SMLP trained with or without
hints. In both SMLP-hints and SMLP-nohints experiments, the neural network was not able 
to generalize or even learn the training set without using standardization in the SMLP intermediate layer,
doing just chance performance. More specifically, in the SMLP-nohints architecture, standardization 
is part of the computational graph, hence the gradients are being backpropagated through it.
The mean and the standard deviation is computed for each hidden unit separately at the
intermediate layer as in Equation \ref{eq:std}. But in order to prevent numerical underflows or
overflows during the backpropagation we have used $\epsilon=1e-8$ (Equation \ref{eq:sigma}).





The benefit of having sparse activations may be specifically important for the ill-conditioned 
problems, for the following reasons. When a hidden unit is ``off'', its gradient (the derivative of
the loss with respect to its output) is usually close to 0 as well, as seen here. 
That means that all off-diagonal
second derivatives involving that hidden unit (e.g. its input weights) are also near 0.
This is basically like removing some columns and rows from the Hessian matrix associated with
a particular example. It has been observed that the condition number of the Hessian
matrix (specifically, its largest eigenvalue) increases as the size of the network 
increases~\citep{Dauphin+Bengio-arxiv-2013}, making training considerably slower
and inefficient~\citep{Dauphin+Bengio-arxiv-2013}. Hence one would expect that as sparsity
of the gradients (obtained because of sparsity of the activations) increases, training
would become more efficient, as if we were training a smaller sub-network for each example,
with shared weights across examples, as in dropouts~\citep{Hinton-et-al-arxiv2012}.

\begin{figure}[htbp!]
 \centering
 \mbox{
 \subfigure[Before standardization.]{
 \includegraphics[width=0.5\linewidth, height=\textheight, keepaspectratio=True]{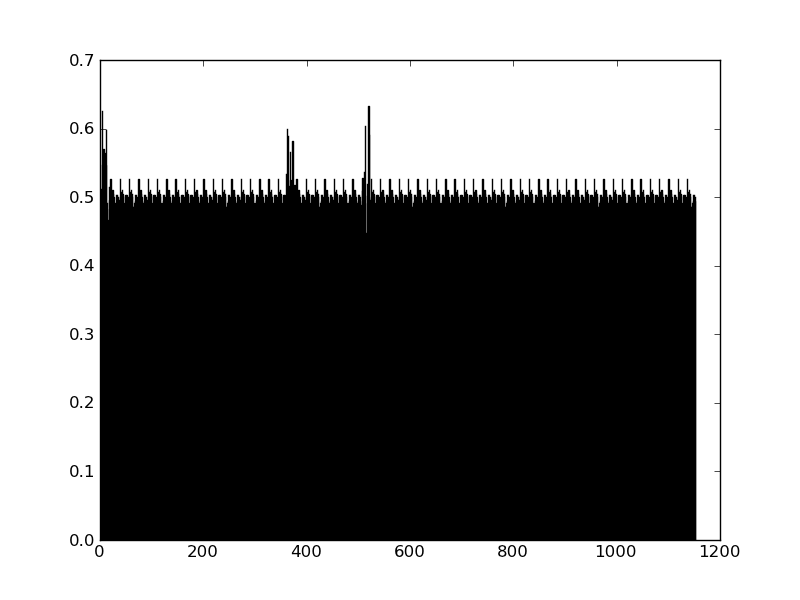}
 \label{fig:nstd_act_bar_nohints}
 }
 \subfigure[After standardization.]{
 \includegraphics[width=0.5\linewidth, height=\textheight, keepaspectratio=True]{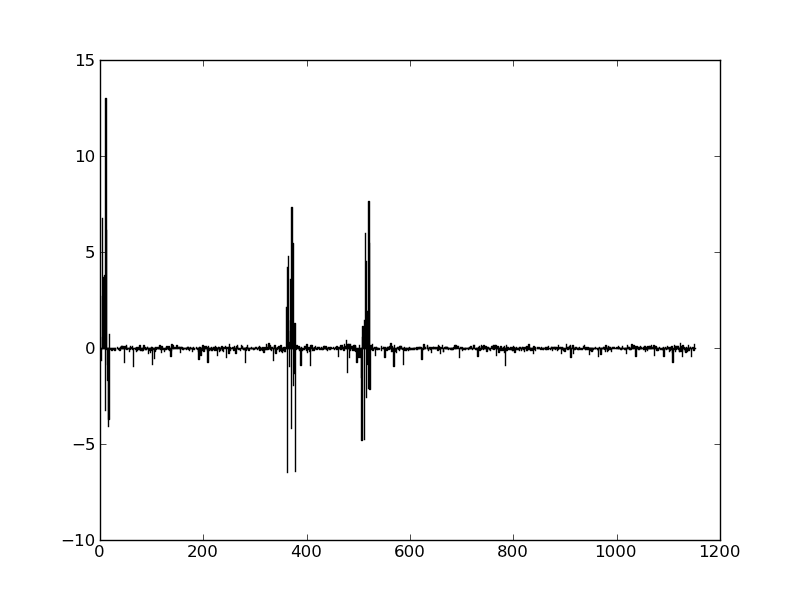}
 \label{fig:std_act_bar_nohints}
 }
 }
\caption{Activations of the intermediate-level hidden units of an SLMP-nohints for 
 a particular examples (x-axis: hidden unit number, y-axis: activation value).
   Left (a): before standardization.
    Right (b): after standardization.}
\label{fig:unstd_std_nohints}
\end{figure}

In Figure \ref{fig:unstd_std_nohints}, the activation of each hidden unit in a bar chart is shown:
the effect of standardization is significant, making the activations sparser.

\begin{figure}[htbp!]
 \centering
 \mbox{
 \subfigure[Before standardization.]{
 \includegraphics[width=0.5\linewidth, height=\textheight, keepaspectratio=True]{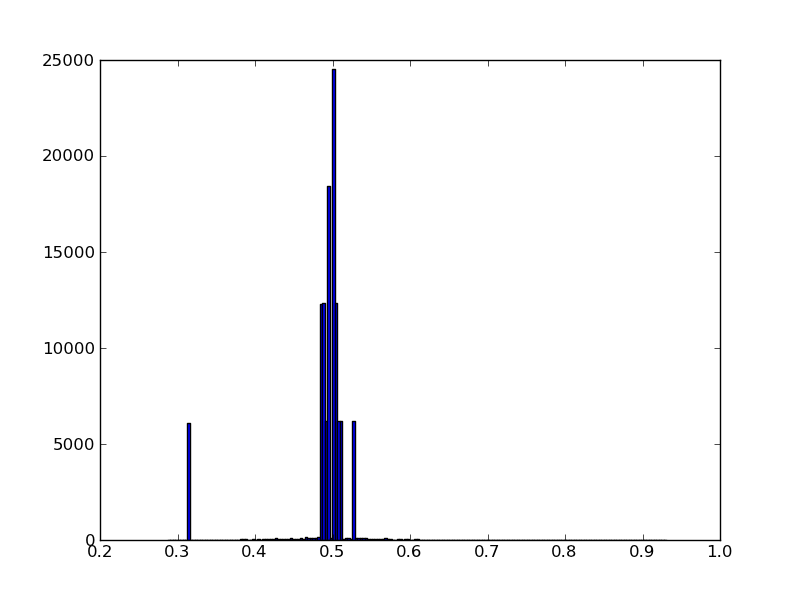}
 \label{fig:unstd_act_hist_nohints}
 }
 \subfigure[After standardization.]{
 \includegraphics[width=0.5\linewidth, height=\textheight, keepaspectratio=True]{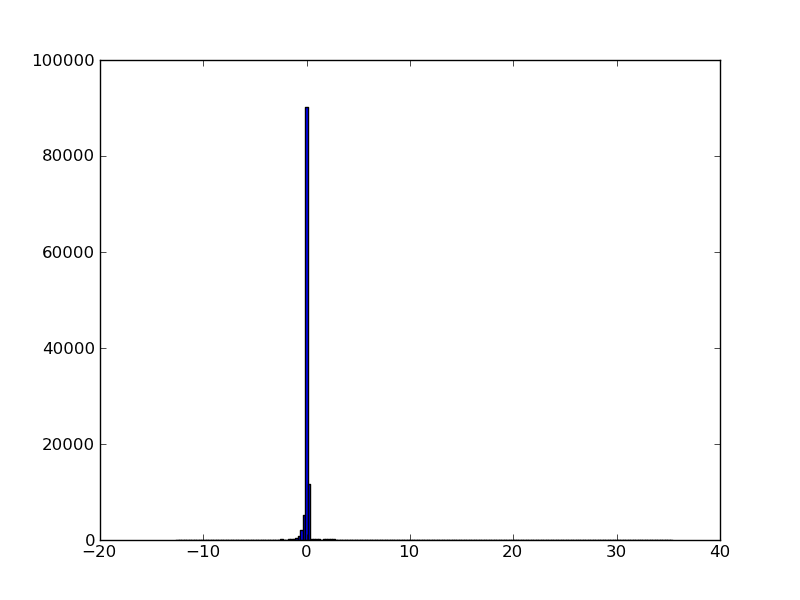}
 \label{fig:std_act_hist_nohints}
 }
}
 \caption{Distribution histogram of activation values of SMLP-nohints intermediate layer.
   Left (a): before standardization.
    Right (b): after standardization.}
\label{fig:unstd_std_acts_hints}
\end{figure}

In Figure \ref{fig:unstd_std_acts_hints}, one can see the activation histogram of the SMLP-nohints
intermediate layer, showing the distribution of activation values, before and after 
standardization. Again the sparsifying effect of standardization is very apparent.

In Figures \ref{fig:unstd_std_acts_hints} and \ref{fig:unstd_std_nohints}, the 
intermediate level activations of
SMLP-nohints are shown before and after standardization. These are
for the same SMLP-nohints architecture whose
results are presented on Table \ref{tab:results_cmp}. 
For that same SMLP,  the Adadelta~\citep{zeiler2012adadelta} adaptive learning
rate scheme has been used, with 512 hidden units for the hidden layer of P1NN and rectifier activation
function. For the output of the P1NN, 11 sigmoidal units have been used while P2NN
had 1200 hidden units with
rectifier activation function. The output nonlinearity of the P2NN is a sigmoid and the 
training objective is the binary crossentropy.

%

\paragraph{Adaptive Learning Rates}
We have experimented with several different adaptive learning rate algorithms. We tried
   rmsprop \footnote{This is learning rate scaling method that is discussed by G. Hinton in his Video Lecture 6.5
       - rmsprop: Divide the gradient by a running average of its recent magnitude. COURSERA: Neural Networks for Machine Learning, 2012.},
   Adadelta \citep{zeiler2012adadelta}, Adagrad~\citep{duchi2010adaptive} 
  and a linearly (1/t) decaying learning rate ~\citep{Bengio-tricks-chapter-2013}. 
   For the SMLP-nohints with sigmoid activation
   function we have found Adadelta\citep{zeiler2012adadelta} converging faster to an effective local minima and usually
   yielding better generalization error compared to the others.



\subsubsection{Deep and Structured MLP with Unsupervised Pre-Training}

Several experiments have been conducted using an architecture similar to the
SMLP-nohints, but by using 
unsupervised pre-training of P1NN, with Denoising Auto-Encoder (DAE) and/or
Contractive Auto-Encoders (CAE).
Supervised fine-tuning proceeds as in the deep and
structured MLP without hints. Because an unsupervised learner may not focus
the representation just on the shapes, a larger number of
intermediate-level units at the output of P1NN has been explored:
previous work on unsupervised pre-training generally
found that larger hidden layers were optimal when using unsupervised
pre-training, because not all unsupervised features will be relevant
to the task at hand. Instead of limiting to 11 units per patch,
we experimented with networks with up to 20 hidden (i.e., code) units 
per patch in the second-layer patch-wise auto-encoder. 

In Appendix \ref{sec:bin_rbm_pento} we also provided the result of some experiments
with binary-binary RBMs trained on 8$\times$8 patches from the 40k training dataset.

In unsupervised pretraining experiments in this paper, both 
\emph{contractive auto-encoder}(CAE) with sigmoid nonlinearity
and binary cross entropy cost function and \emph{denoising
auto-encoder}(DAE) have been used. In the second layer, experiments were performed with
a DAE with rectifier hidden units utilizing L1 sparsity and weight decay on the
weights of the auto-encoder. Greedy layerwise unsupervised training
procedure is used to train the deep auto-encoder architecture \citep{Bengio-nips-2006-small}. In
unsupervised pretraining experiments, tied weights have been used.
Different combinations of CAE and DAE for unsupervised pretraining  have been tested, but none of
the configurations tested managed to learn the Pentomino task, as shown
in Table~\ref{tab:results_cmp}.

\begin{table}[htbp!]
{\tiny%
\newcommand{\mc}[3]{\multicolumn{#1}{#2}{#3}}
\begin{center}
\tiny
\begin{sc}
\begin{tabular}{|l|l|l|l|l|l|l|l|p{2cm}|}
\hline 
\mc{1}{|c|}{\textbf{Algorithm}} & \mc{2}{|c|}{\textbf{20k dataset}} & \mc{2}{|c|}{\textbf{40k dataset}} & \mc{2}{|c|}{\textbf{80k dataset}}\\ \hline
 & \mc{1}{|c|}{\textbf{Training}} & \mc{1}{|c|}{\textbf{Test}} & \mc{1}{|c|}{\textbf{Training}} & \mc{1}{|c|}{\textbf{Test}} & \mc{1}{|c|}{\textbf{Training}} & \mc{1}{|c|}{\textbf{Test}}\\ 
× & \mc{1}{|c|}{\textbf{Error}} & \mc{1}{|c|}{\textbf{Error}} & \mc{1}{|c|}{\textbf{Error}} & \mc{1}{|c|}{\textbf{Error}} & \mc{1}{|c|}{\textbf{Error}} & \mc{1}{|c|}{\textbf{Error}}\\ \hline
\textbf{SVM RBF} & 26.2 & 50.2 & 28.2 & 50.2 & 30.2 & 49.6\\ \hline
\textbf{K Nearest Neighbors} & 24.7 & 50.0 & 25.3 & 49.5 & 25.6 & 49.0\\ \hline
\textbf{Decision Tree} & 5.8 & 48.6 & 6.3 & 49.4 & 6.9 & 49.9\\ \hline
\textbf{Randomized Trees} & 3.2 & 49.8 & 3.4 & 50.5 & 3.5 & 49.1\\ \hline
\textbf{MLP} & 26.5 & 49.3 & 33.2 & 49.9 & 27.2 & 50.1\\ \hline
\textbf{Convnet/Lenet5} & 50.6 & 49.8 & 49.4 & 49.8 & 50.2 & 49.8 \\ \hline
\textbf{Maxout Convnet} & 14.5 & 49.5 & 0.0 & 50.1 & 0.0 & 44.6  \\ \hline
\textbf{2 layer sDA} & 49.4 & 50.3 & 50.2 & 50.3 & 49.7 & 50.3 \\ \hline
\textbf{Struct. Supervised MLP w/o hints} & 0.0 & 48.6 & 0.0 & 36.0 & 0.0 & 12.4 \\ \hline
\textbf{Struct. MLP+CAE Supervised Finetuning} & 50.5  &  49.7 & 49.8  & 49.7 & 50.3 & 49.7 \\ \hline
\textbf{Struct. MLP+CAE+DAE, Supervised Finetuning} & 49.1 & 49.7 & 49.4 & 49.7 & 50.1 & 49.7 \\ \hline
\textbf{Struct. MLP+DAE+DAE, Supervised Finetuning} & 49.5 & 50.3 & 49.7 & 49.8 & 50.3 & 49.7  \\ \hline

\hline\hline
\textbf{Struct. MLP with Hints} & \textbf{0.21} & \textbf{30.7} & \textbf{0} & \textbf{3.1} & \textbf{0} & \textbf{0.01}\\ \hline
\end{tabular}
\end{sc}
\caption{The error percentages with different learning algorithms on Pentomino dataset with different number of training examples.}
\label{tab:results_cmp}
\end{center}
}%
\end{table}

\subsection{Experiments with 1 of K representation}

To explore the effect of changing the complexity of the input representation on the difficulty of the
task, a set of experiments have been designed with symbolic representations of the information in each patch.
In all cases an empty patch is represented with a 0 vector. These representation can be
seen as an alternative input for a P2NN-like network, i.e., they were fed as input to an MLP
or another black-box classifier.

The following four experiments have been conducted, each one using one 
using a different input representation for each patch:

\begin{description}
    \item[Experiment 1-Onehot representation without transformations:] In this experiment several
    trials have been done with a 10-input one-hot vector per patch. Each input corresponds to an object category given in clear, i.e., the ideal input for P2NN if a supervised P1NN perfectly did its job.

    \item[Experiment 2-Disentangled representations:] In this experiment, we did trials with
    16 binary inputs per patch, 10 one-hot bits for representing each object category, 4 
    for rotations and 2 for scaling, i.e., the whole information about the input is given, but
    it is perfectly disentangled. This would be the ideal input for P2NN if an unsupervised P1NN
    perfectly did its job.

    \item[Experiment 3-Onehot representation with transformations:] For each of the ten object
    types there are 8 = 4$\times$2 possible transformations. Two objects in two different patches are the considered ``the same'' (for the final task) if their
    category is the same regardless of the transformations. The one-hot representation of a patch
    corresponds to the cross-product between the 10 object shape classes and the 4$\times$2 
    transformations, i.e., one out
    of 80=10$\times 4 \times 2$ possibilities represented in an 80-bit one-hot vector.
    This also contains all the information about the input image patch, but spread out in
    a kind of non-parametric and non-informative (not disentangled) way, 
    like a perfect memory-based unsupervised
    learner (like clustering) could produce. Nevertheless, the shape class would be easier to
    read out from this representation than from the image representation (it would be an OR
    over 8 of the bits).

    \item[Experiment 4-Onehot representation with 80 choices:] This representation has the same
    1 of 80 one-hot representation per patch but the target task is defined differently. 
    Two objects in two different patches are considered the same iff they have exactly the same 80-bit
    onehot
    representation (i.e., are of the same object category with the same transformation applied).

\end{description}

The first experiment is a sanity check. It was conducted with single hidden-layered
MLP's with rectifier and tanh nonlinearity, and 
the task was learned perfectly (0 error on both training and test dataset) 
with very few training epochs.

\begin{figure}[ht]
 \centering
 \mbox{
 \subfigure[Training and Test Errors for Experiment 4]{
 \includegraphics[width=0.5\linewidth, height=\textheight, keepaspectratio=True]{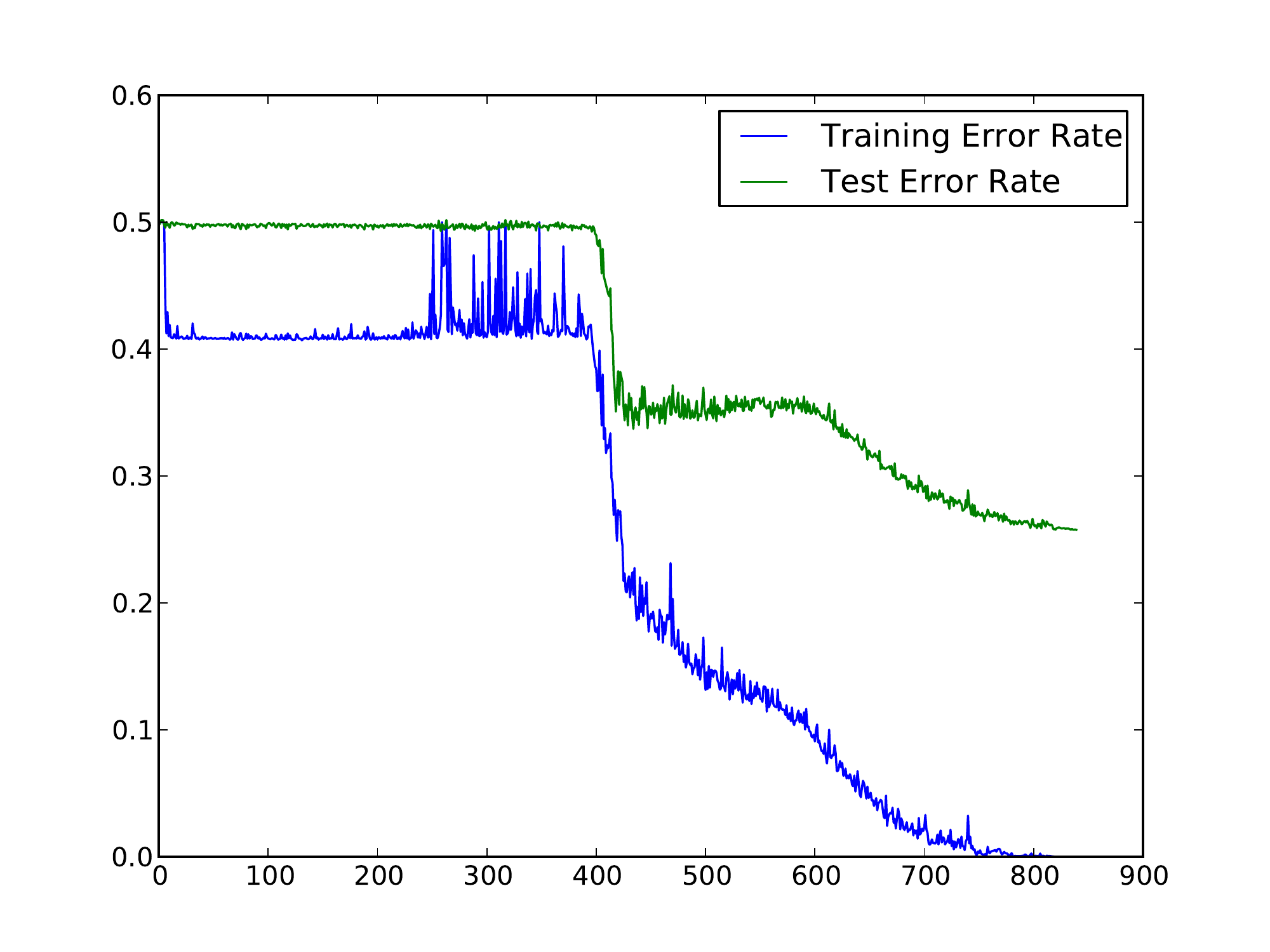}
 \label{fig:80inputs_test_train}
 }
 \subfigure[Training and Test Errors for Experiment 3]{
 \includegraphics[width=0.5\linewidth, height=\textheight, keepaspectratio=True]{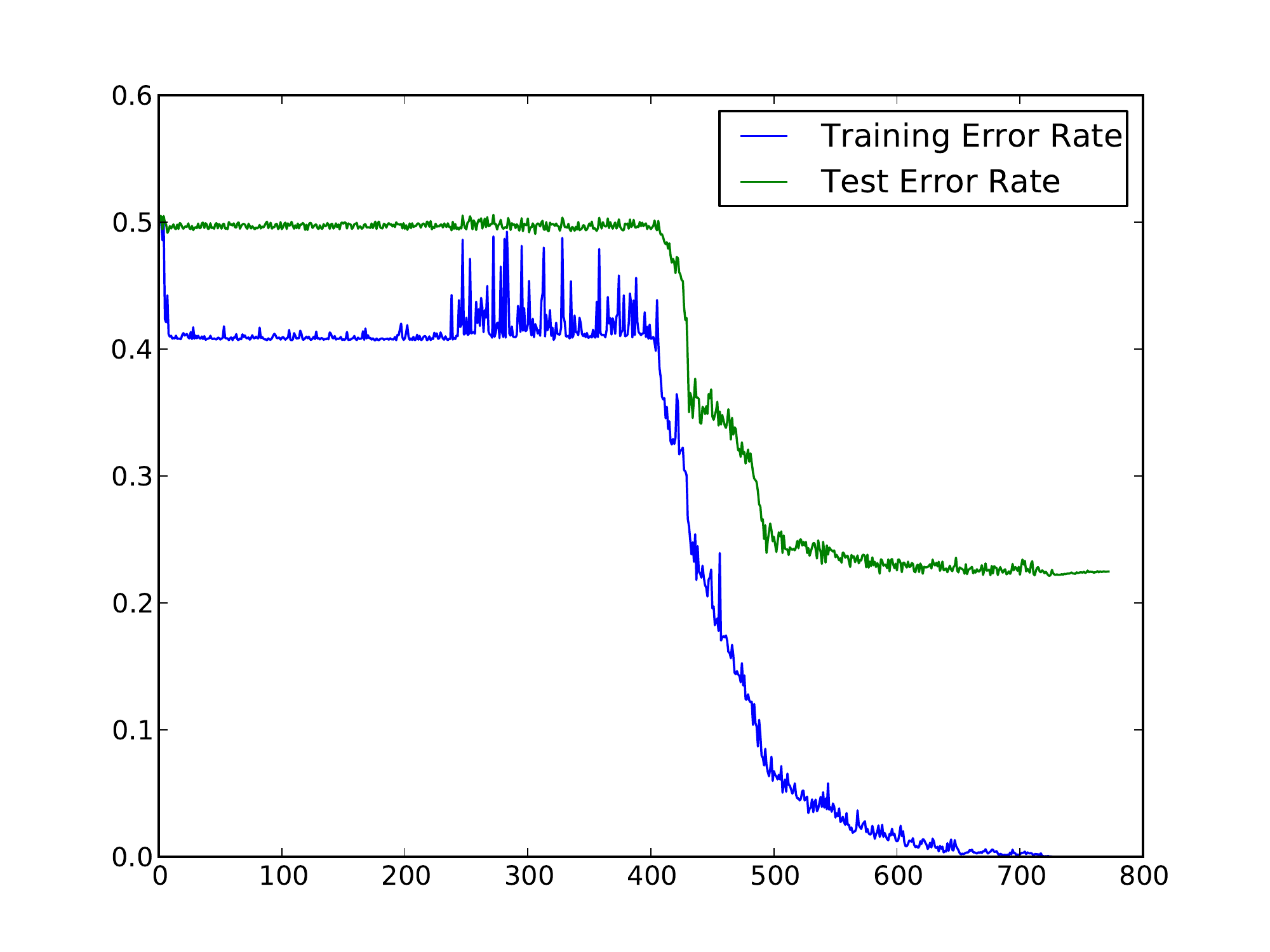}
 \label{fig:80inputs_trans_test_train}
 }
 }
\caption{Tanh MLP training curves. Left (a): The training and test errors of Experiment 3 over 800 training epochs with 100k
    training examples using Tanh MLP.
    Right (b):The training and test errors of Experiment 4 over 700 training epochs with 100k
           training examples using Tanh MLP.}
\label{fig:test_train_err_plots}
\end{figure}

The results of Experiment 2 are given in Table \ref{tab:disentangled}. 
To improve results, we experimented with the
Maxout non-linearity in a feedforward MLP \citep{Goodfellow_maxout_2013} with two hidden layers. 
Unlike the typical Maxout network mentioned in the original paper, regularizers
have been deliberately avoided in order to focus on the optimization
issue, i.e: no weight decay, norm constraint on the weights, or dropout. 
Although learning from a disentangled representation is more difficult than learning from perfect object detectors, it is feasible
with some architectures such as the Maxout network. Note that this representation
is the kind of representation that one could hope an unsupervised learning algorithm
could discover, at best, as argued in~\cite{Bengio+Courville+Vincent-arxiv2012}.

The only results obtained on the validation set for Experiment 3 and Experiment 4 are shown respectively in Table
\ref{tab:80inputs_withtrans} and Table \ref{tab:80inputs}. In these experiments
a tanh MLP with two hidden layers have been tested with the same hyperparameters. In experiment
3 the complexity of the problem comes from the transformations (8=4$\times$2) 
and the number of object types.

But in experiment 4, the only source of complexity of the task comes from the number of different
object types. These results are in between the complete failure and complete success
observed with other experiments, suggesting that the task could become solvable
with better training or more training examples. Figure~\ref{fig:test_train_err_plots} illustrates the progress of training a tanh MLP, on both
the training and test error, for Experiments 3 and 4. Clearly, something has been learned,
but the task is not nailed yet. On experiment 3 for both maxout and tanh the maxout there was a
long plateau where the training error and objective stays almost same. Maxout did just chance on
the experiment for about 120 iterations on the training and the test set. But after 120th
iteration the training and test error started decline and eventually it was able to solve the task.
Moreover as seen from the curves in Figure \ref{fig:test_train_err_plots}\subref{fig:80inputs_test_train}
and \ref{fig:test_train_err_plots}\subref{fig:80inputs_trans_test_train},
the training and test error curves are almost the same for both tasks. This implies that for onehot inputs,
whether you increase the number of possible transformations for each object or 
the number of object categories, as soon as the number of possible configurations
is same, the complexity of the problem is almost the same for the MLP.

\begin{table}[t]
    \small
    \begin{sc}
    \begin{center}
    \begin{tabular}{l|l|l}
        \textbf{Learning Algorithm} & \textbf{Training Error} & \textbf{Test Error} \\ \hline
        SVM                              & 0.0         & 35.6      \\
        Random Forests                   & 1.29        & 40.475    \\
        Tanh MLP                         & 0.0         & 0.0 \\
        Maxout MLP                         & 0.0         & 0.0 \\
    \end{tabular}
    \end{center}
    \end{sc}
    \caption{Performance of different learning algorithms on disentangled representation in Experiment 2.}
    \label{tab:disentangled}
\end{table}

\begin{table}[t]
    \small
    \begin{sc}
    \begin{center}
    \begin{tabular}{l|l|l}
        \textbf{Learning Algorithm} & \textbf{Training Error} & \textbf{Test Error} \\ \hline
        SVM                              & 11.212         & 32.37      \\
        Random Forests                   & 24.839        & 48.915    \\
        Tanh MLP                         & 0.0         & 22.475    \\
        Maxout MLP                       & 0.0         & 0.0 \\
    \end{tabular}
    \end{center}
    \end{sc}
    \caption{Performance of different learning algorithms using a dataset with onehot vector and
        80 inputs as discussed for Experiment 3.}
    \label{tab:80inputs_withtrans}
\end{table}

\begin{table}[t]
    \small
    \begin{sc}
    \begin{center}
    \begin{tabular}{l|l|l}
        \textbf{Learning Algorithm} & \textbf{Training Error} & \textbf{Test Error} \\ \hline
        SVM                              & 4.346          & 40.545      \\ 
        Random Forests                  & 23.456          & 47.345      \\ 
        Tanh MLP                             & 0         & 25.8      \\
    \end{tabular}
    \end{center}
    \end{sc}
    \caption{Performance of different algorithms using a dataset with onehot vector and 80 binary inputs as discussed in Experiment 4.}
    \label{tab:80inputs}
\end{table}

\subsection{Does the Effect Persist with Larger Training Set Sizes?}

The results shown in this section indicate that the problem in the Pentomino task clearly is not
just a regularization problem, but rather basically hinges on 
an optimization problem. Otherwise, we would expect test error to decrease as the
number of training examples increases. This is shown first by studying
the online case and then by studying the ordinary training case with a fixed size
training set but considering increasing training set sizes. In the online minibatch
setting, parameter updates are performed as follows:

\begin{equation}
    \theta_{t+1} = \theta_t - \Delta_{\theta_t} 
\end{equation}

\begin{equation}
    \Delta_{\theta_t} = \epsilon \frac{\sum_i^N \nabla_{\theta_t}L(x_t, \theta_t)}{ N}
\end{equation}

where $L(x_t, \theta_t)$ is the loss incurred on example $x_t$ with parameters
$\theta_t$ , where $t \in \mathcal{Z}^+$ and $\epsilon$ is the learning rate.

Ordinary batch algorithms converge linearly to the optimum $\theta^{*}$, however the noisy gradient
estimates in the online SGD will cause parameter $\theta$ to fluctuate near the local optima.
However, online SGD directly optimizes the expected risk, because the examples are drawn iid from the
ground-truth distribution \citep{bottou2010large}. Thus:

\begin{equation}
    L_{\infty}=E[L(x,\theta)] = \int_x L(x, \theta)p(x)d_x
\end{equation}

where $L_{\infty}$ is the generalization error. Therefore online SGD is trying to minimize the
expected risk with noisy updates. Those noisy updates have the effect of regularizer:

\begin{equation}
    \Delta_{\theta_t} = \epsilon \frac{\sum_i^N \nabla_{\theta_t}L(x_t, \theta_t)}{ N} = \epsilon
\nabla_{\theta_t}L(x,
                                                                                               \theta_t)
    + \epsilon \xi_t
\end{equation}

where $\nabla_{\theta_t}L(x, \theta_t)$ is the true gradient and $\xi_t$ is the zero-mean
stochastic
gradient ``noise'' due to computing the gradient over a finite-size minibatch sample.

We would like to know if the problem with the Pentomino dataset is more a regularization or an optimization
problem. An SMLP-nohints model was trained by online SGD
with the randomly generated online Pentomino stream. 
The learning rate was adaptive, with the Adadelta procedure~\citep{zeiler2012adadelta} on minibatches of 100 examples. 
In the online SGD experiments, two SMLP-nohints that is trained with and without standardization at the
intermediate layer with exactly the same hyperparameters  are tested.
The SMLP-nohints P1NN patch-wise submodel has 2048 hidden units and 
the SMLP intermediate layer has $1152=64\times 18$ hidden units.
The nonlinearity that is used for the intermediate layer is 
the sigmoid. P2NN has 2048 hidden units.

\begin{figure}[htp]
\centering{
\includegraphics[scale=0.6]{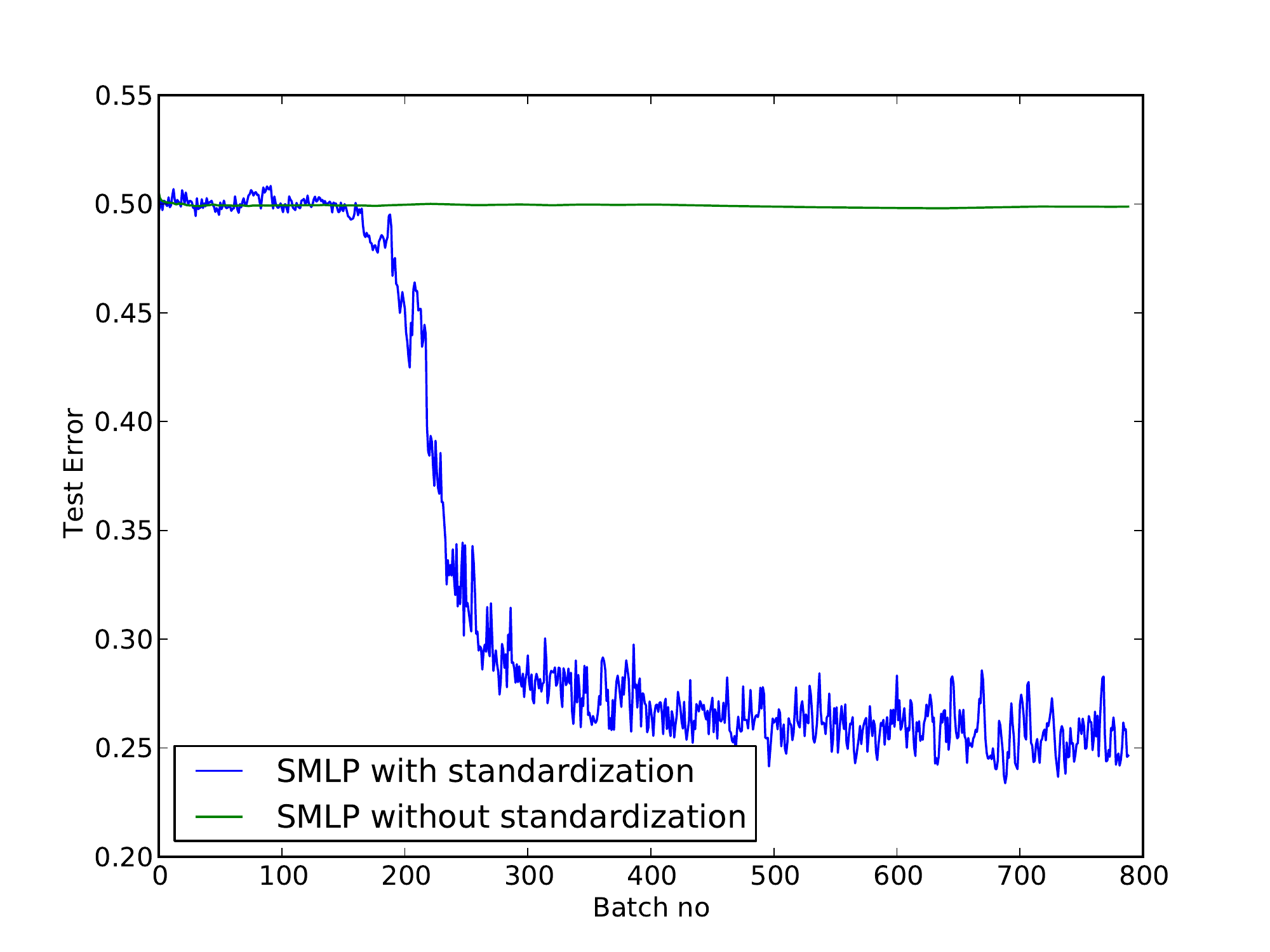}
}
\caption{Test errors of SMLP-nohints with and without standardization in the intermediate layer. Sigmoid as an intermediate 
    layer activation has been used. Each tick (batch no) in the x-axis represents 400 examples.}
\label{fig:std_vs_unstd_test_acc}
\end{figure}

\begin{figure}[htp]
\centering{
\includegraphics[scale=0.6]{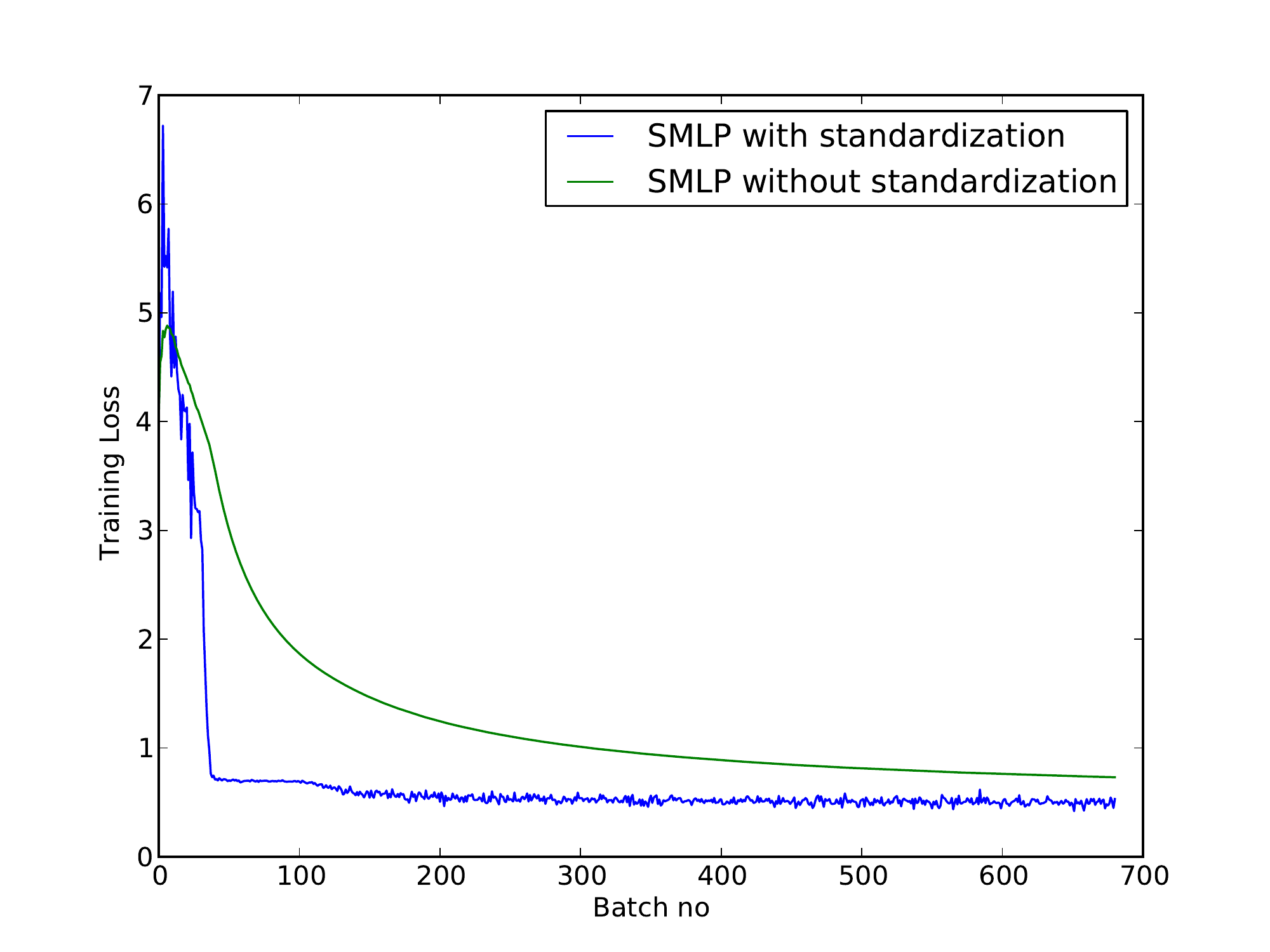}
}
\caption{Training errors of SMLP-nohints with and without standardization in the intermediate
    layer. Sigmoid nonlinearity has been used as an intermediate layer activation function.
        The x-axis is in units of blocks of 400 examples in the training set.}
\label{fig:std_vs_unstd_train_acc}
\end{figure}

SMLP-nohints has been trained either with or without standardization on top of the output units of the P1NN.
The experiments illustrated in Figures 
\ref{fig:std_vs_unstd_test_acc} and \ref{fig:std_vs_unstd_train_acc} are with the same SMLP without hints architecture for which results 
are given in Table \ref{tab:results_cmp}. In those graphs only the results for the training on
the randomly generated 545400 Pentomino samples have been presented. As shown in the plots 
SMLP-nohints was not able to generalize without standardization. Although without 
standardization the training loss seems to decrease initially, it eventually 
gets stuck in a plateau where training loss doesn't change much.

Training of SMLP-nohints online minibatch SGD is performed 
using standardization in the intermediate layer and Adadelta learning rate
adaptation, on 1046000 training examples
from the randomly generated Pentomino stream. At the end of the training, 
test error is down to 27.5\%, which is much better than chance but
from from the score obtained with SMLP-hints of near 0 error.


In another SMLP-nohints experiment {\em without standardization} 
the model is trained with the 1580000 Pentomino
examples using online minibatch SGD. P1NN has 2048 hidden units and 16 sigmoidal outputs per patch.
for the P1NN hidden layer. P2NN has 1024 hidden units for the hidden layer. Adadelta is used to adapt 
the learning rate. At the end of training this SMLP, the test error remained
stuck, at 50.1\%.




\subsubsection{Experiments with Increased Training Set Size}

\begin{figure}[htbp!]
 \centering
 \vspace*{-1mm}
 \includegraphics[scale=0.5]{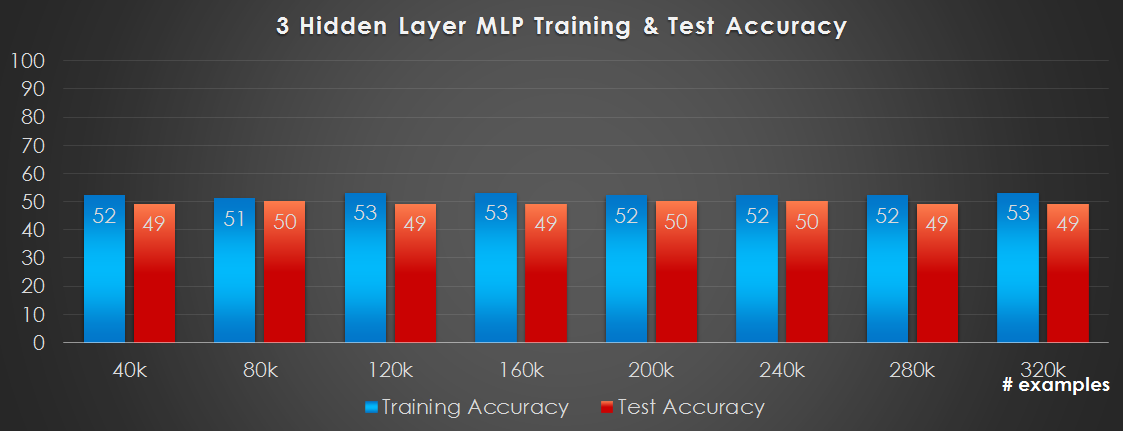}
 \vspace*{-2mm}
 \caption{Training and test error bar charts for a regular MLP with 3 hidden layers.
     There is no significant improvement on the generalization error of the MLP 
     as the new training examples are introduced.}
 \label{fig:error_bars}
\end{figure}

Here we consider the effect of training different learners with different numbers of training examples.
For the experimental results shown in Table \ref{tab:results_cmp},
3 training set sizes (20k, 40k and 80k examples) had been used. Each dataset was generated with different
random seeds (so they do not overlap). Figure
\ref{fig:error_bars} also shows the error bars for an ordinary MLP
with three hidden layers, for a larger range of training set sizes,
between 40k and 320k examples. The number of training epochs
is 8 (more did not help), and there are three hidden layers with 2048 feature detectors.
The learning rate we used in our experiments is 0.01. The activation function of the MLP is
a tanh nonlinearity, while the L1, L2 penalty coefficients are both 1e-6.


Table \ref{tab:results_cmp} shows that, without guiding hints, none of
the state-of-art learning algorithms could perform noticeably better than a random
predictor on the test set.  This shows the importance of intermediate hints
introduced in the SMLP. The decision trees and SVMs can overfit the training
set but they could not generalize on the test set. Note that the numbers reported
in the table are for hyper-parameters selected based on validation set error,
hence lower training errors are possible if avoiding all regularization and taking
large enough models. On the training set, the MLP with two large hidden
layers (several thousands) could reach nearly 0\% training error, but still
did not manage to achieve good test error.

In the experiment results shown in Figure~\ref{fig:error_bars}, we evaluate
the impact of adding more training data for the fully-connected MLP. As mentioned before
for these experiments we have used a MLP with three hidden layers where each layer
has 2048 hidden units. The tanh$(\cdot)$ activation function is used with 0.05 learning rate
and minibatches of size 200.

As can be seen from the figure, adding more training examples did not help
either training or test error (both are near 50\%, with training error slightly
lower and test error slightly higher), 
reinforcing the hypothesis that the difficult encountered 
is one of optimization, not of regularization.

\subsection{Experiments on Effect of Initializing with Hints}

\begin{figure}[htp]
\centering{
\includegraphics[scale=0.6]{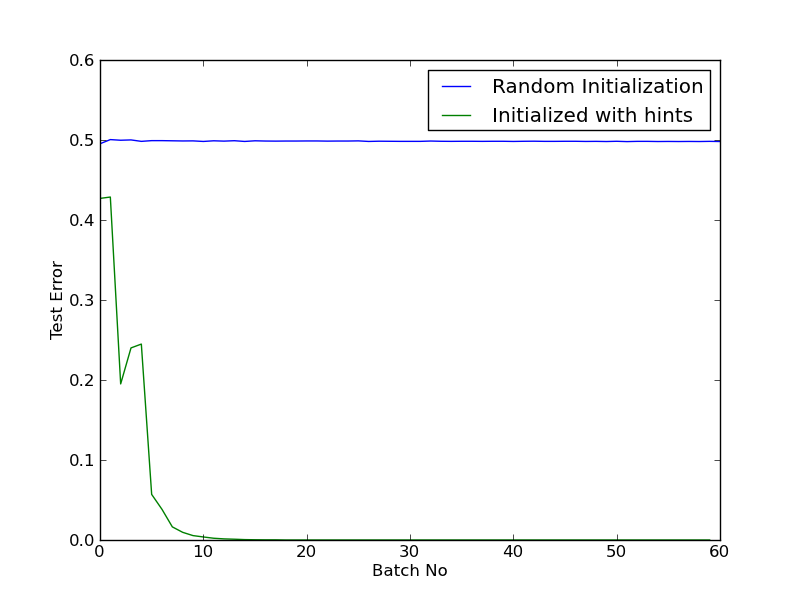}
}
\caption{Plots showing the test error of SMLP with random initialization vs initializing with hint based training.}
\label{fig:opt_hint_init}
\end{figure}

Initialization of the parameters in a neural network can have a big impact on the learning and
generalization \citep{GlorotAISTATS2010}. Previously \cite{Erhan+al-2010} showed that
initializing the parameters of a neural network with unsupervised pretraining guides the
learning towards basins of attraction of local minima that provides 
better generalization from the training dataset. In this section we
analyze the effect of initializing the SMLP with hints and then continuing without hints at the
rest of the training.
For experimental analysis of hints based initialization, SMLP is trained for 1 training epoch using the hints and for 60
epochs it is trained without hints on the 40k examples training set. We also compared the same architecture
with the same hyperparameters, against to SMLP-nohints trained for 61 iterations on the same dataset.
After one iteration of hint-based training SMLP obtained 9\% training error and 39\% test error.
Following the hint based training, SMLP is trained without hints for 60 epochs, but at epoch 18,
it already got 0\% training and 0\% test error. The hyperparameters for this experiment and the
experiment that the results shown for the SMLP-hints in Table \ref{tab:results_cmp} are 
the same. The test results
for initialization with and without hints are shown on Figure \ref{fig:opt_hint_init}. This figure
suggests that initializing with hints can give the same generalization performance but training takes
longer.

\subsubsection{Further Experiments on Optimization for Pentomino Dataset}

With extensive hyperparameter optimization and using standardization in the intermediate level of
the SMLP with softmax nonlinearity, SMLP-nohints was able to get 5.3\% training and 6.7\% test error
on the 80k Pentomino training dataset. We used the 2050 hidden units for the hidden
layer of P1NN and 11 softmax output per patch. For the P2NN, we used 1024 hidden units with
sigmoid and learning rate 0.1 without using any adaptive learning rate method. This SMLP uses 
a rectifier nonlinearity
for hidden layers of both P1NN and P2NN. Considering that architecture uses softmax as the intermediate activation function of SMLP-nohints. It is
very likely that P1NN is trying to learn the presence of specific Pentomino shape in a given patch. This architecture 
has a very large capacity in the P1NN, that probably provides it enough capacity to learn the presence
of Pentomino shapes at each patch effortlessly.

An MLP with 2 hidden layers, each 1024 rectifier units, was trained
using LBFGS (the implementation from the scipy.optimize library) 
on 40k training examples, with gradients computed on batches of 10000 examples at each iteration.
However, after convergence of training, the MLP was still doing chance on the test dataset.

We also observed that using linear units for the intermediate layer yields better generalization
error without standardization compared to using activation functions such as sigmoid, tanh and RELU
for the intermediate layer. SMLP-nohints was able to get 25\% generalization error with linear
units without standardization whereas all the other activation functions that has been tested failed to
generalize with the same number of training iterations without standardization and hints. This suggests that
using non-linear intermediate-level activation functions 
without standardization introduces an optimization
difficulty for the SMLP-nohints, maybe because the intermediate level acts like
a bottleneck in this architecture.

\section{Conclusion and Discussion}

In this paper we have shown an example of task which seems almost
impossible to solve by standard black-box machine learning algorithms,
but can be almost perfectly solved when one encourages a semantics for the intermediate-level
representation that is guided by prior knowledge. The
task has the particularity that it is defined by the composition
of two non-linear sub-tasks (object detection on one hand, and
a non-linear logical operation similar to XOR on the other hand).

What is interesting is that in the case of the neural network, we can
compare two networks with exactly the same architecture but a different
pre-training, one of which uses the known intermediate concepts to teach an
intermediate representation to the network. With enough capacity and training time they can
{\em overfit} but did not not capture the essence of the task, as seen by
test set performance. 

We know that a structured deep network can learn the task, if it is initialized in the right place,
and do it from very few training examples. Furthermore we have shown that if 
one pre-trains SMLP with
hints for only one epoch, it can nail the task. But the exactly same architecture which started training from random
initialization, failed to generalize.

Consider the fact that even SMLP-nohints with standardization after being trained
using online SGD on 1046000 generated examples and still gets 27.5\% test error. This is an
indication that the problem is not a regularization problem but possibly an inability to
find a {\em good effective local minima of generalization error}.

What we hypothesize is that for most initializations and architectures (in particular the fully-connected ones),
although it is possible to find a {\em good effective local minimum of training error}
when enough capacity is provided, it is difficult (without the proper initialization)
to find a good local minimum of generalization error. On the other hand, when the network architecture
is constrained enough but still allows it to represent a good solution
(such as the structured MLP of our experiments), it seems that the optimization
problem can still be difficult and even training error remains stuck high if the standardization
isn't used. Standardization obviously makes the training objective of the SMLP easier to optimize and helps it
to find at least a {\em better effective local minimum of training error}. This finding suggests that by using specific
architectural constraints and sometimes domain specific knowledge about the problem, one can alleviate
the optimization difficulty that generic neural network architectures face.

It could be that the combination of the network architecture and training procedure
produces a training dynamics that tends to yield into these minima that are
poor from the point of view of generalization error, even when they manage to nail training
error by providing enough capacity. Of course, as the number of examples increases, we would expect this
discrepancy to decrease, but then the optimization problem could still make
the task unfeasible in practice.  Note however that our preliminary
experiments with increasing the training set size (8-fold) for MLPs did not reveal
signs of potential improvements in test error yet, as shown in
Figure~\ref{fig:error_bars}. Even using online training on 545400 Pentomino examples,
the SMLP-nohints architecture was still doing far from perfect in terms of generalization error
(Figure~\ref{fig:std_vs_unstd_test_acc}).

These findings bring supporting evidence to the
``Guided Learning Hypothesis'' and ``Deeper
Harder Hypothesis'' from~\citet{Bengio-chapter-2013}:
higher level abstractions, which are expressed by composing
simpler concepts, are more difficult to learn (with the learner
often getting in an effective local minimum ), but that difficulty
can be overcome if another agent provides hints of the importance
of learning other, intermediate-level abstractions which are relevant
to the task.

Many interesting questions remain open. Would a network without any
guiding hint eventually find the solution with a enough training time
and/or with alternate parametrizations? To what extent is ill-conditioning a core issue?
The results with LBFGS were disappointing but changes in the architectures (such
as standardization of the intermediate level) seem to make training much easier.
Clearly, one can reach good solutions from an
appropriate initialization, pointing in the direction of an issue with local minima,
but it may be that good solutions are also reachable from
other initializations, albeit going through a tortuous ill-conditioned
path in parameter space. Why did our attempts at learning the intermediate
concepts in an unsupervised way fail? Are these results specific to the
task we are testing or a limitation of the unsupervised feature learning algorithm
tested? Trying with many more unsupervised variants and exploring
explanatory hypotheses for the observed failures could help us answer
that. Finally, and most ambitious, can we solve these kinds of problems
if we allow a community of learners to collaborate and collectively
discover and combine partial solutions in order to obtain solutions
to more abstract tasks like the one presented here? Indeed, we would
like to discover learning algorithms that can solve such tasks without
the use of prior knowledge as specific and strong as the one used
in the SMLP here. These experiments could be inspired by and inform
us about potential mechanisms for collective learning through cultural
evolutions in human societies.

\acks{We would like to thank to the ICLR 2013 reviewers for their insightful comments,
and NSERC, CIFAR, Compute Canada and Canada Research Chairs for funding.}

\bibliography{cultrefs,strings,ml,aigaion}

\section{Appendix}

\subsection{Binary-Binary RBMs on Pentomino Dataset}
\label{sec:bin_rbm_pento}
We trained binary-binary RBMs (both visible and hidden are binary)
on 8$\times$8 patches extracted from the Pentomino Dataset using PCD (stochastic
maximum likelihood), a
weight decay of .0001 and a sparsity penalty\footnote{implemented as TorontoSparsity in pylearn2, see the
yaml file in the repository for more details}. 
We used 256 hidden units and trained by SGD with a batch size of 32 and
a annealing learning rate \citep{Bengio-tricks-chapter-2013} starting from 1e-3 with annealing 
rate 1.000015.
The RBM is trained with momentum starting from 0.5. 
The biases are initialized to -2 in
order to get a sparse representation. 
The RBM is trained for 120 epochs (approximately 50 million updates).

After pretraining the RBM, its parameters are used to initialize
the first layer of an SMLP-nohints network.
As in the usual architecture of the SMLP-nohints on top of P1NN, there is an intermediate layer.
Both P1NN and the intermediate layer have a sigmoid nonlinearity, and the intermediate
layer has 11 units per location. This SMLP-nohints is trained with Adadelta and standardization at the
intermediate layer \footnote{In our auto-encoder experiments we directly fed features
to P2NN without standardization and Adadelta.}.

\begin{figure}[htp]
\centering{
\includegraphics[scale=0.6]{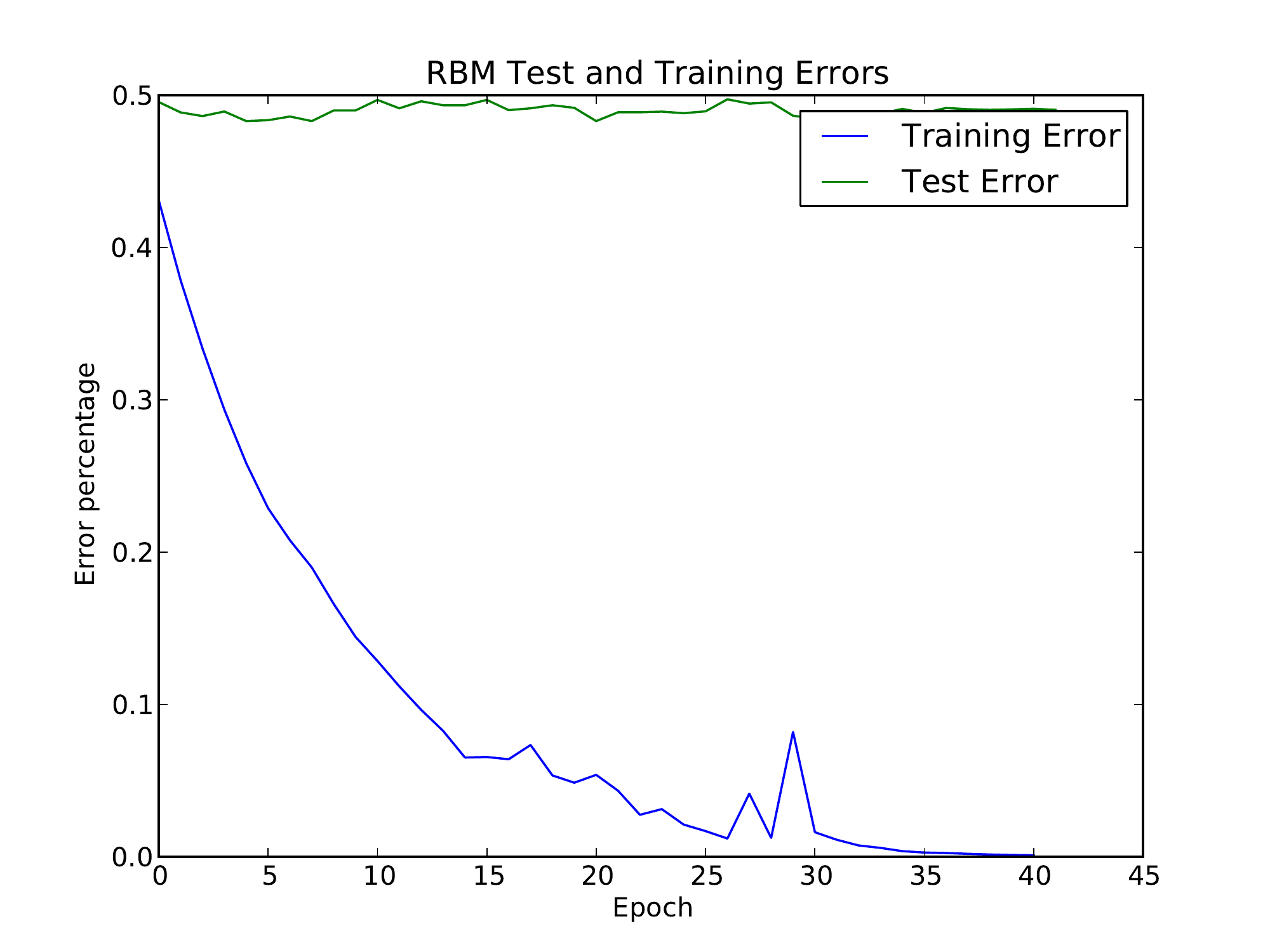}
}
\caption{Training and test errors of an SMLP-nohints network whose first layer
is pre-trained as an RBM. Training error
    reduces to 0\% at epoch 42, but test error is still chance.}
\label{fig:rbm_pre_errors}
\end{figure}

\begin{figure}[htp]
\centering{
\includegraphics[scale=0.8]{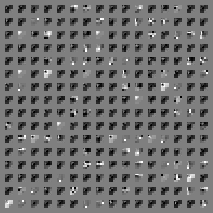}
}
\caption{Filters learned by the binary-binary RBM after training on the 40k examples. 
The RBM did learn the edge structure of Pentomino shapes.}
\label{fig:rbm_pento_filters}
\end{figure}

\begin{figure}[htp]
\centering{
\includegraphics[scale=1.0]{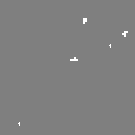}
}
\caption{100 samples generated from trained RBM. All the generated samples are 
valid Pentomino shapes.}
\label{fig:rbm_pento_samples}
\end{figure}

\subsection{Experimental Setup and Hyper-parameters}
\label{sec:setup}

\subsubsection{Decision Trees}

We used the decision tree implementation in the
scikit-learn~\citep{pedregosa2011scikit} python package which is an
implementation of the CART (Regression Trees) algorithm. The CART algorithm
constructs the decision tree recursively and partitions the input space such that
the samples belonging to the same category are grouped together
\citep{olshen1984classification}. We used The Gini index as the impurity
criteria. We evaluated the hyper-parameter configurations with a grid-search. We
cross-validated the maximum depth ($max\_depth$) of the tree (for
preventing the algorithm to severely overfit the training set) and
minimum number of samples required  to create a split ($min\_split$).
20 different configurations of hyper-parameter values were evaluated. 
We obtained the best validation error with $max\_depth=300$ and
$min\_split=8$.

\subsubsection{Support Vector Machines}

We used the ``Support Vector Classifier (SVC)'' implementation from the
scikit-learn package which in turn uses the libsvm's Support Vector
Machine (SVM) implementation. Kernel-based SVMs are non-parametric models that
map the data into a high dimensional space and 
separate different classes with hyperplane(s) such that the support
vectors for each category will be separated by a large margin. We
cross-validated three hyper-parameters of the model using grid-search: $C$,
$\gamma$ and the type of kernel($kernel\_type$). $C$ is the penalty term 
(weight decay) for the SVM and $\gamma$ is a hyper-parameter that controls the width of the
Gaussian for the RBF kernel. For the polynomial kernel, $\gamma$ controls
the flexibility of the classifier (degree of the polynomial)
as the number of parameters increases \citep{hsu2003practical, ben2010user}. 
We evaluated forty-two hyper-parameter configurations. That includes, two kernel types:
$\{RBF,~Polynomial\}$; three gammas: $\{1e-2,~1e-3,~1e-4\}$ for the RBF
kernel, $\{1,2,5\}$ for the polynomial kernel, and seven $C$ values among:
$\{0.1, 1, 2, 4, 8, 10, 16\}$. As a result of the grid search and
cross-validation, we have obtained the best test error by using the RBF kernel,
with $C=2$ and $\gamma=1$.
 
\subsubsection{Multi Layer Perceptron}

We have our own implementation of Multi Layer Perceptron based on the 
Theano~\citep{bergstra+al:2010-scipy-short} machine learning libraries. We have
selected 2 hidden layers, the rectifier activation function, and
2048 hidden units per layer. We cross-validated three hyper-parameters of the 
model using random-search, sampling the learning rates $\epsilon$ in log-domain, and 
selecting $L1$ and $L2$ regularization penalty coefficients in sets of fixed values,
evaluating 64 hyperparameter values. The range of the hyperparameter 
values are $\epsilon \in [0.0001, 1]$, $L1 \in \{0., 1e-6, 1e-5, 1e-4\}$ and 
$L2 \in \{0, 1e-6, 1e-5\}$.
As a result, the following were selected: $L1=1e-6$, $L2=1e-5$ and $\epsilon=0.05$.

\subsubsection{Random Forests}

We used scikit-learn's implementation of ``Random Forests'' decision
tree learning. The Random Forests algorithm creates an ensemble of
decision trees by randomly selecting for each tree a subset of features 
and applying bagging to combine the individual decision trees \citep{Breiman01}.  We have used grid-search and
cross-validated the $max\_depth$, $min\_split$, and number of trees ($n\_estimators$). We have
done the grid-search on the following hyperparameter values, $n\_estimators\in\{5, 10, 15, 25, 50\}$, 
$max\_depth\in \{100, 300, 600, 900\}$, and $min\_splits\in\{1, 4, 16\}$. We obtained the best validation error
with $max\_depth=300$, $min\_split=4$ and $n\_estimators=10$.

\subsubsection{k-Nearest Neighbors}

We used scikit-learn's implementation of k-Nearest Neighbors (k-NN). k-NN
is an instance-based, lazy learning algorithm that selects the training
examples closest in Euclidean distance to the input query.
It assigns a class label to the test example based on the
categories of the $k$ closest neighbors. The hyper-parameters we have
evaluated in the cross-validation are the number of neighbors ($k$) and
$weights$. The $weights$ hyper-parameter can be either ``uniform'' or
``distance''. With ``uniform'', the value assigned to the query point 
is computed by the majority vote of the nearest neighbors. With ``distance'',
each value assigned to the query point is computed by weighted majority votes
where the weights are computed with the inverse distance between the query
point and the neighbors. We have used $n\_neighbours\in\{1, 2, 4, 6, 8, 12\}$ and
$weights\in\{"uniform", "distance"\}$ for hyper-parameter search. As a result of 
cross-validation and grid search, we obtained the best validation
error with $k=2$ and $weights$=``uniform''.

\subsubsection{Convolutional Neural Nets}

We used a Theano~\citep{bergstra+al:2010-scipy-short} implementation of
Convolutional Neural Networks (CNN) from the deep learning tutorial at
\url{deeplearning.net}, which is based on a vanilla version of a CNN
\citet{lecun1998gradient}. Our CNN has two convolutional layers. Following
each convolutional layer, we have a max-pooling layer. On top of the
convolution-pooling-convolution-pooling layers there is an MLP with one
hidden layer. In the
cross-validation we have sampled 36 learning rates in log-domain in the
range $[0.0001, 1]$ and the number of filters from the range $[10, 20, 30, 40,
  50, 60]$ uniformly. For the first convolutional layer we used 9$\times$9
receptive fields in order to guarantee that each object fits inside the
receptive field.  As a result of random hyperparameter search and doing manual
hyperparameter search on the validation dataset, the following values were selected:
\begin{itemize}
    \item The number of features used for the first layer is 30 and the second layer is 60.
    \item For the second convolutional layer, 7$\times$7 receptive fields.
    The stride for both convolutional layers is 1.
    \item Convolved images are downsampled by a factor of 2$\times$2 at each pooling operation.
    \item The learning rate for CNN is 0.01 and it was trained for 8 epochs.
\end{itemize}

\subsubsection{Maxout Convolutional Neural Nets}

We used the pylearn2 (\url{https://github.com/lisa-lab/pylearn2}) 
implementation of maxout convolutional networks~\citep{Goodfellow_maxout_2013}.
There are two convolutional layers in the selected architecture, 
without any pooling. In the last convolutional
layer, there is a maxout non-linearity. 
The following were selected by cross-validation: learning rate, number of channels for the
both convolution layers, number of kernels for the second layer and number of units and pieces
per maxout unit in the last layer, a linearly decaying learning rate, momentum
starting from 0.5 and saturating to 0.8 at the 200'th epoch. 
Random search for the hyperparameters was used to evaluate 48 different hyperparameter configurations
on the validation dataset. For the first convolutional layer, 8$\times$8 kernels were selected 
to make sure that each Pentomino shape fits into the kernel. Early stopping was used and 
test error on the model that has the best validation error is reported. Using norm constraint
on the fan-in of the final softmax units yields slightly better result on the validation dataset.

As a result of cross-validation and manually tuning the hyperparameters we used the following
hyperparameters:

\begin{itemize}
    \item 16 channels per convolutional layer. 600 hidden units for the maxout layer.
    \item 6x6 kernels for the second convolutional layer.
    \item 5 pieces for the convolution layers and 4 pieces for the maxout layer per maxout units.
    \item We decayed the learning rate by the factor of 0.001 and the initial learning rate is
    0.026367. But we scaled the learning rate of the second convolutional layer by a constant
    factor of 0.6.
    \item The norm constraint (on the incoming weights of each unit) is 1.9365.
\end{itemize}

\begin{figure}[htp]
\centering{
\includegraphics[scale=1.0]{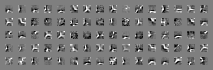}
}
\caption{Maxout convolutional net first layer filters. 
Most of the filters were able to learn the basic edge structure of the Pentomino shapes.}
\label{fig:maxconvnet_filters}
\end{figure}

Figure \ref{fig:maxconvnet_filters} shows the first layer filters of the maxout convolutional net, 
after being trained on the 80k training set for 85 epochs.

\subsubsection{Stacked Denoising Auto-Encoders}
Denoising Auto-Encoders (DAE) are a form of regularized 
auto-encoder~\citep{Bengio-Courville-Vincent-TPAMI2013}. The DAE forces the hidden layer
 to discover more robust features and prevents it from simply learning the identity by 
reconstructing the input from a corrupted version of it~\citep{Vincent-JMLR-2010}. Two DAEs
were stacked, resulting in an unsupervised transformation with two hidden
layers of 1024 units each. Parameters of all layers are then fine-tuned with supervised fine-tuning using logistic regression 
as the classifier and 
SGD as the gradient-based optimization algorithm. The stochastic corruption process
is binomial (0 or 1 replacing each input value, with probability 0.2).
The selected learning rate 
is $\epsilon_0=0.01$ for the DAe and $\epsilon_1=0.1$ 
for supervised fine-tuning. Both L1 and L2 penalty for the DAEs and 
for the logistic regression layer are set to 1e-6.

\paragraph{CAE+MLP with Supervised Finetuning:}

A regularized auto-encoder which sometimes outperforms the DAE is the Contractive
Auto-Encoder (CAE), \citep{Rifai-icml2012}, which penalizes the Frobenius norm 
of the Jacobian matrix of derivatives of the hidden units with respect to the CAE's
inputs. The CAE serves as pre-training for an MLP, and in the supervised fine-tuning state,
the Adagrad method was used to automatically tune the learning rate \citep{duchi2010adaptive}.

After training a CAE with 100 sigmoidal units patch-wise, the features extracted on each patch
are concatenated and fed as input to an MLP. The selected Jacobian penalty coefficient is 2,
the learning rate for pre-training is 0.082 with batch size of 200 and 
200 epochs of unsupervised learning are performed on the training set. For supervised finetuning,
the learning rate is 0.12 over 100 epochs, L1 and L2 regularization penalty terms respectively are 1e-4 and 1e-6,
and the top-level MLP has 6400
hidden units.




\paragraph{Greedy Layerwise CAE+DAE Supervised Finetuning:}

For this experiment we stack a CAE with sigmoid non-linearities
and then a DAE with rectifier non-linearities
during the pre-training phase.
As recommended by \citet{Glorot+al-AI-2011} we have used a softplus nonlinearity for reconstruction,
$softplus(x)=log(1+e^x)$.
We used an L1 penalty on the rectifier outputs
to obtain a sparser representation with rectifier non-linearity and L2 regularization 
to keep the non-zero weights small.

The main difference between the DAE and CAE is that the DAE yields more robust reconstruction
whereas the CAE obtains more robust features \citep{Rifai+al-2011-small}.

As seen on Figure \ref{fig:net_arch_nohints} the weights U and V are shared on each patch
and we concatenate the outputs of the last auto-encoder on each patch to feed it as an input to
an MLP with a large hidden layer.

We used 400 hidden units for the CAE and 100 hidden units for DAE. The learning rate used for 
the CAE is
0.82 and for DAE it is 9*1e-3. The corruption level for the DAE (binomial noise) 
is 0.25 and the contraction level for the CAE is 2.0. The L1
regularization penalty for the DAE is 2.25*1e-4 and the L2 penalty is 9.5*1e-5. 
For the supervised finetuning
phase the learning rate used is 4*1e-4 with L1 and L2 penalties respectively 1e-5 and 1e-6. 
The top-level MLP has  6400 hidden units. The auto-encoders are each trained for 150 epochs
while the whole MLP is fine-tuned for 50 epochs.

\paragraph{Greedy Layerwise DAE+DAE Supervised Finetuning:} 
For this architecture, we have trained two layers of denoising auto-encoders greedily 
and performed supervised finetuning after unsupervised pre-training. The motivation for using two 
denoising auto-encoders is the fact that rectifier nonlinearities work well with the
deep networks but it is difficult to train CAEs with the rectifier non-linearity. 
We have used the same type of denoising auto-encoder that is used for the
greedy layerwise CAE+DAE supervised finetuning experiment.

In this experiment we have used 400 hidden units for the first layer DAE and 100
hidden units for the second layer DAE. The other hyperparameters for DAE and supervised
finetuning are the same as with the {\emph{CAE+DAE MLP Supervised Finetuning}} experiment.

\end{document}